\documentclass{article} 
\usepackage{iclr2027_conference,times}
\usepackage{amsmath}
\usepackage{amssymb}
\usepackage{amsthm}
\usepackage{graphicx}
\usepackage{booktabs}
\usepackage{multirow}
\usepackage{array}
\usepackage{enumitem}

\usepackage{algorithm}
\usepackage{algpseudocode}

\theoremstyle{definition}

\usepackage{amsmath,amsfonts,bm}

\def\eqref#1{equation~\ref{#1}}

\def\1{\bm{1}}

\DeclareMathAlphabet{\mathsfit}{\encodingdefault}{\sfdefault}{m}{sl}
\SetMathAlphabet{\mathsfit}{bold}{\encodingdefault}{\sfdefault}{bx}{n}

\def\eqref#1{(\ref{#1})}
\newcommand{\numDeltaCtenSSD}{0.29}  
\newcommand{\numParamsCtenSSD}{3.63}  
\newcommand{\numDeltaCtenLFSSD}{0.11}  
\newcommand{\numParamsCtenLFSSD}{3.09}  
\newcommand{\numDfCtenUnAct}{0.00}  
\newcommand{\numDeltaCtenUnAct}{0.08}  
\newcommand{\numParamsCtenUnAct}{4.80}  
\newcommand{\numOrigAccCten}{94.97}  
\newcommand{\numParamsCtwentySSD}{0.60}  
\newcommand{\numDeltaCtwentyLFSSD}{0.24}  
\newcommand{\numParamsCtwentyLFSSD}{0.79}  
\newcommand{\numDeltaCtwentyUnAct}{0.80}  
\newcommand{\numParamsCtwentyUnAct}{4.95}  
\newcommand{\numOrigAccCtwenty}{85.34}  
\newcommand{\numParamsChundredSSD}{0.23}  
\newcommand{\numDeltaChundredLFSSD}{0.38}  
\newcommand{\numParamsChundredLFSSD}{0.12}  
\newcommand{\numDeltaChundredUnAct}{1.15}  
\newcommand{\numParamsChundredUnAct}{2.18}  
\newcommand{\numDrGapUnActChundred}{1.15}  
\newcommand{\numOrigAccChundred}{77.35}  
\newcommand{\numVitDeltaSSD}{33.74}  
\newcommand{\numVitWorstDrSSD}{9.57}  
\newcommand{\numVitDeltaExclThreeSSD}{35.71}  
\newcommand{\numVitDeltaUnAct}{11.49}  
\newcommand{\numVitWorstDrUnAct}{85.33}  
\newcommand{\numVitCloserUnAct}{3}  
\newcommand{\numVitDeltaExclThreeUnAct}{13.60}  
\newcommand{\numNstarSSDCten}{5000}  
\newcommand{\numNstarLFSSDCten}{100}  
\newcommand{\numNstarUnActCten}{25}  
\newcommand{\numNstarSSDCtwenty}{2500}  
\newcommand{\numNstarLFSSDCtwenty}{25}  
\newcommand{\numNstarUnActCtwenty}{5}  
\newcommand{\numNstarSSDChundred}{500}  
\newcommand{\numNstarLFSSDChundred}{5}  
\newcommand{\numNstarUnActChundred}{5}  
\newcommand{\numDeltaSSDCtenNfive}{89.85}  
\newcommand{\numMinDrSSDCten}{29.60}  
\newcommand{\numDeltaLFSSDCtenNfive}{67.07}  
\newcommand{\numMinDrLFSSDCten}{11.26}  
\newcommand{\numDeltaUnActCtenNfive}{0.21}  
\newcommand{\numMinDrUnActCten}{95.51}  
\newcommand{\numDeltaUnActCtwentyNfive}{0.77}  
\newcommand{\numDeltaLFSSDChundredNfive}{0.44}  
\newcommand{\numDeltaUnActChundredNfive}{1.37}  
\newcommand{\numWorstDropSSDCten}{85.34}  
\newcommand{\numWorstDropLFSSDCten}{86.09}  
\newcommand{\numUnActRunsEthree}{125}  
\newcommand{\numTimeUnActCtenKone}{0.58}  
\newcommand{\numSpeedupAmortCtenKone}{3.3}  
\newcommand{\numSpeedupColdCtenKone}{35}  
\newcommand{\numTimeUnActCtwentyKone}{0.38}  
\newcommand{\numSpeedupAmortCtwentyKone}{3.0}  
\newcommand{\numTimeUnActChundredKone}{0.12}  
\newcommand{\numSpeedupAmortChundredKone}{1.8}  
\newcommand{\numSpeedupColdChundredKone}{215}  
\newcommand{\numKoneMinusSelectedCten}{0.06}  
\newcommand{\numKoneMinusSelectedCtwenty}{0.05}  
\newcommand{\numKoneMinusSelectedChundred}{0.00}  
\newcommand{\numSimultaneousSSDCtenKfive}{94.54}  
\newcommand{\numSimultaneousUnActCtenKfive}{84.76}  
\newcommand{\numSimultaneousSSDCtwentyKfive}{89.97}  
\newcommand{\numSimultaneousUnActCtwentyKfive}{60.51}  
\newcommand{\numSimultaneousSSDChundredKfive}{81.29}  
\newcommand{\numSimultaneousUnActChundredKfive}{43.66}  
\newcommand{\numSequentialGoldDrCtenKfive}{97.92}  
\newcommand{\numSeqDrUnActCtenRone}{95.2}  
\newcommand{\numSeqDfSSDCtenRtwo}{47.7}  
\newcommand{\numSeqDfSSDCtenRfive}{76.1}  
\newcommand{\numSeqDrUnActCtenRfive}{96.6}  
\newcommand{\numSeqDrUnActCtwentyRone}{85.1}  
\newcommand{\numSeqDfSSDCtwentyRtwo}{32.5}  
\newcommand{\numSeqDrUnActCtwentyRthree}{83.7}  
\newcommand{\numSeqDfSSDCtwentyRfive}{62.1}  
\newcommand{\numSeqDrSSDCtwentyRfive}{85.7}  
\newcommand{\numSeqDrUnActCtwentyRfive}{70.1}  
\newcommand{\numSeqDrWorstUnActCtwentyRfive}{59.6}  
\newcommand{\numSeqDrUnActChundredRone}{76.4}  
\newcommand{\numSeqDfSSDChundredRfive}{0.5}  
\newcommand{\numSeqDrSSDChundredRfive}{77.1}  
\newcommand{\numSeqDrUnActChundredRfive}{70.8}  
\newcommand{\numVitVariantFfnDlaLasttwo}{44.24}  
\newcommand{\numVitVariantFfnPosLastfour}{95.41}  
\newcommand{\numRelearnGoldDrStepfiftyCtenMone}{40.6}  
\newcommand{\numRelearnGoldDrStepfiftyChundredMone}{4.4}  
\newcommand{\numRelearnGoldDrStepfiftyCtwentyMone}{35.8}  
\newcommand{\numRatioSSDUnActCtenNfive}{430}  
\newcommand{\numRatioLFSSDUnActCtenNfive}{321}  
\newcommand{\numRatioSSDUnActCtwentyNfive}{94}  
\newcommand{\numRatioLFSSDUnActCtwentyNfive}{10}  
\newcommand{\numWorstDropBaselinesMax}{86}  
\newcommand{\numWorstDropUnActMax}{2.45}  
\newcommand{\numMIAMaxUnlearn}{0.9}  
\newcommand{\numSeqDfMaxUnActCtenCtwenty}{2.7}  
\newcommand{\numSeqDfMinOrderSSDCtwentyRfive}{51.4}  
\newcommand{\numSeqDfMaxOrderSSDCtwentyRfive}{68.9}  
\newcommand{\numSeqDrPerRequestUnActChundred}{1.4}  
\newcommand{\numVitDrGapUnAct}{6.5}  
\newcommand{\numVitRatioSSDUnAct}{2.9}  
\newcommand{\numSelRatioCten}{4.38}  
\newcommand{\numSelRatioCtwenty}{3.19}  
\newcommand{\numSelRatioChundred}{2.54}  
\newcommand{\numCoverageRuns}{1500}  
\newcommand{\numCoverageExceedNaive}{489}  
\newcommand{\numCoverageExceedKmN}{0}  
\newcommand{\numLabelFreeSSDDiffMinM}{10.9}  
\newcommand{\numLabelFreeSSDDiffMaxM}{11.0}  
\newcommand{\numVitRelearnDfGoldMfiveSfive}{0.0}  
\newcommand{\numVitRelearnDfSSDMfiveSfive}{33.6}  
\newcommand{\numVitRelearnDfUnActMfiveSfive}{87.3}  
\newcommand{\numVitFsDeltaUnActNfive}{12.63}  
\newcommand{\numVitFsWorstDropUnAct}{13.7}  
\newcommand{\numVitFsDeltaSSDNfivehundred}{88.78}  
\newcommand{\numVitFsWorstDropSSD}{88.5}  
\newcommand{\numVitFsTimeUnActFull}{73}  
\newcommand{\numVitFsTimeSSDColdFull}{1,373}  
\newcommand{\numVitFsTimeSSDImportance}{1,248}  
\newcommand{\numVitFsSpeedupCold}{19}  
\newcommand{\numVitFsSpeedupAmort}{1.7}  
\newcommand{\numVitFsTimeUnActNfive}{0.1}  

\newcommand{\numDepthDeltaMaskCten}{0.09}  
\newcommand{\numDepthAgrFMaskCten}{62.4}  
\newcommand{\numDepthAgrFUnActCten}{31.4}  
\newcommand{\numDepthAgrFSSDCten}{65.5}  
\newcommand{\numDepthAgrFLFSSDCten}{62.5}  
\newcommand{\numDepthDfUnActOutCten}{0.6}  
\newcommand{\numDepthDeltaMaskCtwenty}{0.25}  
\newcommand{\numDepthDfUnActOutCtwenty}{20.2}  
\newcommand{\numDepthDeltaMaskChundred}{0.33}  
\newcommand{\numDepthDfUnActOutChundred}{41.4}  
\newcommand{\numDepthProbeMin}{0.98}  
\newcommand{\numDepthProbeGoldMin}{0.96}  
\newcommand{\numDepthDfUnActInMin}{20.6}  
\newcommand{\numDepthDfUnActInMax}{51.7}  
\newcommand{\numRfSameCount}{8}  
\newcommand{\numRfCells}{9}  
\newcommand{\numRfMaxGap}{0.03}  
\newcommand{\numLodoMaxSSD}{32.42}  
\newcommand{\numLodoMaxLFSSD}{93.72}  
\newcommand{\numLodoMaxUnAct}{3.46}  
\newcommand{\numSingleCfgUnAct}{(99, 0.01, 20)}  
\newcommand{\numOracleSSDNfiveCten}{87.5}  
\newcommand{\numOracleSSDNfiveCtwenty}{64.6}  
\newcommand{\numOracleSSDNfiveChundred}{46.1}  
\newcommand{\numOracleSSDSmallMin}{25}  

\usepackage{hyperref}
\usepackage{url}

\title{UnAct: Gradient-Free \textbf{Un}learning via Targeted \textbf{Act}ivation Intervention}

\author{Saeed Abdul Muizz\thanks{Equal contribution.}, Aayat Rafiq\footnotemark[1], Iqra Altaf Gillani \& Janibul Bashir \\
Gaash Lab \\
National Institute of Technology Srinagar \\
Srinagar, Jammu and Kashmir, India \\
\texttt{abdul\_2024bmec088@nitsri.ac.in}, \texttt{aayat\_2024bite015@nitsri.ac.in} \\
\texttt{\{iqraaltaf, janibbashir\}@nitsri.ac.in}
}
\iclrfinalcopy 
\begin{document}

\maketitle

\begin{abstract}
Machine unlearning seeks to remove the influence of designated training data from a trained model without retraining from scratch. Retrain-free methods such as Selective Synaptic Dampening (SSD) and its label-free variant LFSSD avoid full retraining but still require backpropagation and parameter importance computed over the entire dataset. We ask: \emph{what happens when a deletion request arrives with only a few images of the class to be forgotten?} To answer this question, we introduce \textbf{UnAct}, a gradient-free class-unlearning method that needs only forward passes over the forget images. UnAct scores late-layer units by their responses, attenuates the most responsive connections, and repeats this for up to 20 rounds using no gradients, no labels, and no retained data. On ResNet-18 trained with CIFAR-10, CIFAR-20, and CIFAR-100, UnAct is competitive with SSD and LFSSD when forgetting entire classes and, unlike them, never collapses the network when forget data is scarce. On ResNet-18, across all tested sizes, UnAct's retain accuracy stays within 2.5 points of retraining, while SSD and LFSSD, at their full-class operating points, lose up to 86 points on some classes. With five forget images on CIFAR-10, UnAct's distance to retraining is 0.21 points, against 67 for LFSSD and 90 for SSD, and re-selecting SSD's threshold at each size with an oracle does not close the gap. In preliminary transfer to ViT-B/16, UnAct's distance to retraining is 11.5 against 33.7 for SSD, and a request is 19$\times$ faster than SSD when SSD computes its importance at request time. The code is available at \url{https://github.com/abdulmuizz0903/UnAct}.

\end{abstract}

\section{Introduction}

A trained model can retain traces of the examples it has seen. Simply deleting a record from the dataset does not erase its influence; the model may continue to encode sensitive information, leaving it exposed to attacks such as membership inference or model inversion~\citep{shokri2017membership,fredrikson2015model}. This challenge has sparked the field of machine unlearning, which aims to selectively remove the impact of designated training data while retaining useful knowledge from the rest~\citep{nguyen2025survey}. Beyond legal imperatives like the GDPR’s right to erasure~\citep{voigt2017eu}, unlearning is crucial for practical reasons: models must be able to discard mislabeled, corrupted, outdated, or harmful data without incurring the prohibitive cost of retraining from scratch, especially when deletion requests are frequent or affect only a small fraction of the dataset.

Existing unlearning methods reduce this cost in different ways. Some retain
information from the original training process so that the contribution of
deleted data can later be reversed, while others perform additional optimization after training~\citep{graves2021amnesiac,chundawat2023can}. A
particularly attractive alternative is \textit{post-hoc, retrain-free
unlearning}, which directly modifies a trained model without retraining on the
retained data. Selective Synaptic Dampening
(SSD)~\citep{foster2024fast} is a representative of
this approach: it uses Fisher information to estimate how important each
parameter is to the forget data and to the full training set, and dampens the
parameters that are disproportionately important to the data to be forgotten.
Its label-free variant, LFSSD~\citep{foster2024loss},
replaces the Fisher information with the gradient of the model's squared output norm, so that no labels are needed. Both methods still calculate the gradients, and both require an importance estimate over the full training set, which can be computed once and stored. We argue that the model's current activations provide a more direct signal for targeted unlearning: activations show which internal units the model actually uses to respond to the data to be forgotten.

Based on this insight, we introduce \emph{UnAct}, a forward-only method for post-hoc class-level unlearning built on four design choices. (i)~A \emph{classifier-relative scope}: UnAct edits only the late units that write to the representation the classifier reads. (ii)~A \emph{forget-only score} of how much signal each unit carries from the forget images to the classifier, computed from forward passes with no labels, gradients or retained data; on transformers this requires measuring a unit's contribution to the class logit rather than its activation. (iii)~\emph{Rank-based selection}, which fixes the size of every edit in advance, independently of how many forget images arrive or how the scores are scaled. (iv)~\emph{Iterative soft attenuation} of each selected unit's incoming and outgoing weights with re-scoring, so that later rounds reach units that earlier rounds missed instead of deepening the same edit. Activation-based pruning \citep{hu2016network} also ranks units by activation, but to remove the ones a network uses least, once, for compression; UnAct differs in all four respects, and (iii) is what bounds its edit when only a handful of forget images arrive, the regime in which importance-ratio dampening can collapse the network.

We evaluate UnAct against SSD and LFSSD on ResNet-18 across CIFAR-10, CIFAR-20, and CIFAR-100, and against SSD on ViT-B/16. With the full forget set on ResNet-18, UnAct reduces forget accuracy to 0\% while keeping retain accuracy within \numDrGapUnActChundred{} points of retraining on every dataset, competitive with both baselines. With few forget images the methods separate: across all forget-set sizes, UnAct's retain accuracy on ResNet-18 stays within \numWorstDropUnActMax{} points of retraining, whereas SSD and LFSSD lose up to \numWorstDropBaselinesMax{} points on some classes, and with five images on CIFAR-10 UnAct's distance to retraining is \numDeltaUnActCtenNfive{} points, against \numDeltaLFSSDCtenNfive{} for LFSSD and \numDeltaSSDCtenNfive{} for SSD. A one-round configuration (not the selected 20-round one on CIFAR-10 and CIFAR-20) removes a class in under 0.6\,s, \numSpeedupAmortChundredKone--\numSpeedupAmortCtenKone$\times$ faster than SSD with its importance precomputed. On ViT-B/16, UnAct's distance to retraining is \numVitDeltaUnAct{} against \numVitDeltaSSD{} for SSD, and a request is \numVitFsSpeedupCold$\times$ faster than SSD from cold. As with existing methods, UnAct improves output-level unlearning without removing the class from the learned representation. Our contributions can be summarized as follows:

\begin{enumerate}
    \item We introduce \textbf{UnAct}, a forward-only method for post-hoc class-level unlearning that reads no gradients, labels or retained data, and whose rank-based selection bounds the edit independently of the number of forget images.

    \item We establish the effectiveness and efficiency of UnAct on convolutional networks, with a preliminary transfer to transformers. UnAct is competitive with SSD and LFSSD on full forget sets, remains robust where the baselines collapse, requires far fewer forget examples, and, like the baselines, admits operating-point selection without a retrained model.

    \item We show that iterative activation re-ranking widens the set of attenuated units rather than repeatedly deepening the same edit. The resulting coverage of the layer organizes UnAct's hyperparameters, and the number of rounds trades request time against closeness to retraining (Section~\ref{sec:attenuation}, Appendix~\ref{app:coverage}).

    \item We analyze what the edit changes inside the network. Units selected by activation alone become increasingly class-selective as the threshold rises (Appendix~\ref{app:selectivity}), and on ResNet-18 editing only the last residual block suffices to forget (Section~\ref{sec:scope}, Table~\ref{tab:main}). We also calibrate the standard evaluation: a trivial logit mask is as close to retraining as any method, and no method, including ours, removes the class from the penultimate features (Section~\ref{sec:findings}, Appendix~\ref{app:depth}).
\end{enumerate}

\section{Related Work}
\label{sec:related}

Machine unlearning methods span a spectrum defined by how much of the original training process they revisit at forgetting time. \emph{Retraining-based approaches} retrain the model fully or partially, \emph{optimization-based approaches} fine-tune the trained model, and \emph{retrain-free approaches} directly modify the parameters associated with the forget set. Below, we review these approaches, together with methods that remove knowledge by editing individual units, and contrast them with UnAct.

\begin{table}[!t]
\caption{Summarized Comparison of unlearning families upon deletion request}
\label{tab:compare}
\centering
\small
\renewcommand{\arraystretch}{1.15}
\begin{tabular}{>{\raggedright\arraybackslash}p{3.5cm}>{\raggedright\arraybackslash}p{4.8cm}>{\raggedright\arraybackslash}p{4.3cm}}
\toprule
\textbf{Method} & \textbf{Requirements} & \textbf{Limitations} \\
\midrule
\emph{Retraining-based}~\citep{ginart2019making,bourtoule2021machine,graves2021amnesiac} & Retained data and full or partial retraining; SISA and Amnesiac also need shards, checkpoints or logged updates & Compute and storage cost; SISA and Amnesiac must be adopted at training time \\
\midrule[0.2pt]
\emph{Optimization-based}~\citep{chundawat2023can,kurmanji2023towards,jia2023model,fan2024salun} & Labels and/or retained data, many gradient steps & Fine-tuning cost, data access at request time \\
\midrule[0.2pt]
\emph{SSD}~\citep{foster2024fast} & Labels, backward passes, Fisher information over the full dataset & Approximate method, unreliable with few forget images \\
\midrule[0.2pt]
\emph{LFSSD}~\citep{foster2024loss} & Backward passes, sensitivity over the full dataset & Approximate method \\
\midrule[0.2pt]
\emph{Unit pruning}~\citep{wang2022federated,pochinkov2024dissecting,chang2024zero} & Retain or reference data, often fine-tuning & Data access, permanent removal of units \\
\midrule
\emph{UnAct (ours)} & Forward passes over the forget images & Approximate method \\
\bottomrule
\end{tabular}
\end{table}

\paragraph{Retraining-based unlearning.} Retraining the model from scratch on the retained data guarantees complete removal of the forget set, making it the oracle baseline~\citep{ginart2019making,cao2015towards,li2025machine}. However, it is computationally prohibitive, introduces latency for each deletion request, and requires permanent storage of the retained data. SISA~\citep{bourtoule2021machine} reduces this cost by splitting the data into shards, training on each separately, and retraining only the shard that contains the deleted samples. Amnesiac unlearning~\citep{graves2021amnesiac} instead logs the updates made during training and subtracts those that involved the forget samples. Both methods must be set up at training time and require storing checkpoints or updates. SISA loses accuracy when deletions are frequent or shards are small~\citep{bourtoule2021machine}, and Amnesiac unlearning can damage retained classes when forget and retain samples appear in the same batches~\citep{graves2021amnesiac}. UnAct, by contrast, is applied to an already-trained model and needs no retained data, checkpoints, or logs. Unlike retraining, it is an approximate method, so we use retrained models as the reference throughout our evaluation.

\paragraph{Optimization-based unlearning.} Most approximate methods fine-tune the trained model, for example with gradient ascent on the forget set, fine-tuning on the retained data, teacher--student objectives such as Bad Teacher~\citep{chundawat2023can} and SCRUB~\citep{kurmanji2023towards}, or sparsity-regularized fine-tuning~\citep{jia2023model}. SalUn~\citep{fan2024salun} computes a weight saliency map from the gradient of the forgetting loss and fine-tunes only the most salient weights, relabeling forget samples randomly while training on the retained data. These methods need labels, the retained data, or both, as well as many optimization steps. UnAct targets the setting in which neither labels nor retained data are available when a deletion request arrives. We therefore compare it with methods that work under the same conditions, and we make no claim of superiority over optimization-based methods, which use more information. Several methods relax data access but still optimize with gradients: Boundary Unlearning~\citep{chen2023boundary} and JiT~\citep{foster2024zero} use only forget samples, zero-shot unlearning~\citep{chundawat2023zero} synthesizes proxy data, and few-shot unlearning~\citep{yoon2022few} recovers a proxy training set by model inversion. UnAct shares their goal of working from few forget samples but performs no optimization.

\paragraph{Retrain-free dampening.} Selective Synaptic Dampening (SSD)~\citep{foster2024fast} computes the diagonal Fisher information of every parameter over the forget set and over the full training set, and dampens the parameters that are disproportionately important for the forget set. SSD is fast and retrain-free, but it needs labels and a backward pass for every request, and its authors note that it is sensitive to its hyperparameters. Loss-Free SSD (LFSSD)~\citep{foster2024loss} removes the need for labels by replacing the Fisher information with a label-free sensitivity measure, the gradient of the squared output norm~\citep{aljundi2018memory}, but it still requires backward passes and sensitivity computed over the full dataset. ASSD~\citep{schoepf2024parameter} selects $\alpha$ adaptively. Because none of these methods retrains the model, SSD and LFSSD are our main baselines, and LFSSD, which also reads no labels, is the closest to UnAct. UnAct instead uses only the forward activations of the forget images, and, as we show in Section~\ref{sec:data-efficiency}, it remains reliable when only a few forget images are available, where SSD does not.

\paragraph{Unit-level removal.} A separate line of work removes knowledge by editing individual units rather than individual weights. Removing a small set of units in a scene classifier removes its ability to recognize particular classes~\citep{bau2020understanding}. For unlearning, \citet{wang2022federated} score channels by a TF-IDF measure of class discrimination, prune the most discriminative channels for the target class in a federated model, and then fine-tune; \citet{pochinkov2024dissecting} prune language-model neurons that are much more active on the forget data than on the retained data; \citet{chang2024zero} perturb neurons ranked by layer-wise relevance propagation; and knowledge-neuron methods edit the MLP neurons attributed to a fact~\citep{dai2022knowledge}. Activation- and importance-based pruning criteria for model compression~\citep{hu2016network,molchanov2017pruning,li2017pruning} follow a similar idea for efficiency, removing the units a network uses least. UnAct is closest to these methods but differs in four ways: it scores units from forward passes on the forget images alone, with no retained data and no relevance backward pass; it selects a fixed fraction of units by rank, so the size of its edit does not depend on the scores or on the number of forget images; it weakens units over several rounds with re-scoring instead of removing them at once; and it needs no fine-tuning afterwards.

\paragraph{Positioning of UnAct.} UnAct combines the efficiency of retrain-free methods with a unit-level edit whose size is fixed in advance. However, like SSD and LFSSD, UnAct is an approximate method, and we measure its forgetting empirically against retrained models. Since accuracy-based metrics and simple membership attacks can overstate forgetting~\citep{lynch2024eight,hayes2024inexact,carlini2022membership}, we also examine how deeply its edit removes the forget class (Appendix~\ref{app:depth}). Table~\ref{tab:compare} summarizes the comparison of UnAct with the existing methods.

\section{Method}
\label{sec:method}

\begin{figure}[t]
\begin{center}
\includegraphics[width=\linewidth]{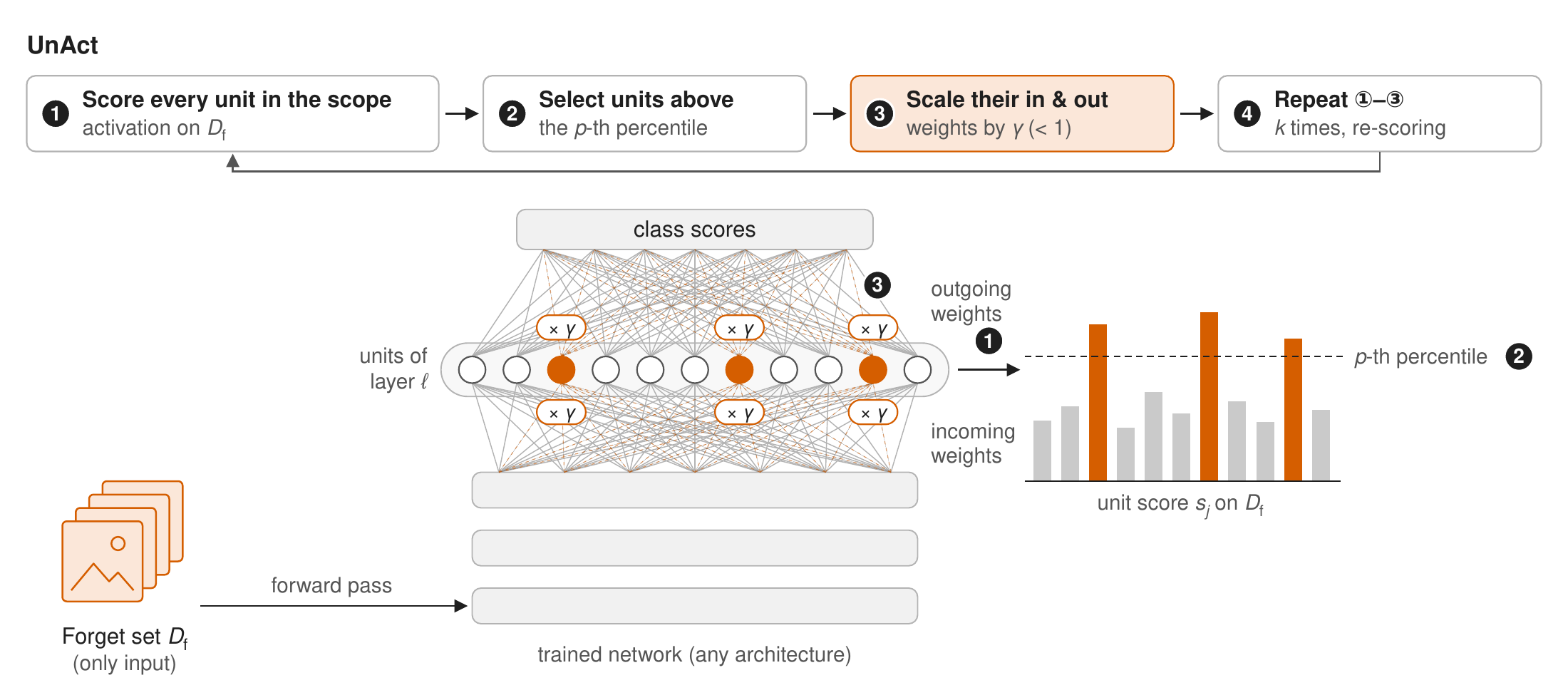}
\end{center}
\caption{\textbf{Overview of UnAct.} Forward passes on the forget set
$\mathcal{D}_f$ score every unit of the target layer (1). Units above the
$p$-th percentile are selected (2), and their incoming and outgoing weights are
multiplied by $\gamma$ (3). The round is repeated $k$ times, re-scoring the
edited model each time (4). No labels, backward passes or retained data are
used.}
\label{fig:method}
\end{figure}

UnAct edits a trained classifier by weakening the internal units that carry the forget class to its output. Each round (Figure~\ref{fig:method}) scores the units of a classifier-relative scope by forward passes on the forget images, selects a fixed fraction of them by rank, and attenuates their incoming and outgoing parameters; the edited model is then rescored. Unlike activation-based pruning \citep{hu2016network}, which removes weakly used units once to compress a model, UnAct targets the units the forget class uses most, never removes them, and re-ranks after every edit.

\subsection{Problem Formulation}
\label{sec:setup-method}

Let $f_\theta$ be a classifier trained on a dataset $\mathcal{D}$. Given a target class $c$, let $\mathcal{D}_f$ be the available forget images of class $c$, either the whole class or a subset of it, and let $\mathcal{D}_r$ be the remaining training data. Class-level unlearning seeks parameters $\theta'$ with which the model no longer retains class $c$ but behaves as before on the other classes. A model retrained from scratch on $\mathcal{D}_r$ is the reference for ideal unlearning. UnAct is post-hoc and uses only the $n$ images in $\mathcal{D}_f$: it does not access $\mathcal{D}_r$, read labels, optimize a loss, or run a backward pass.

\subsection{Scope and Units}
\label{sec:scope}

A \emph{unit} is an internal feature with its own incoming parameters $\mathrm{In}_j$, which compute it, and outgoing parameters $\mathrm{Out}_j$, through which it reaches later layers, such as a convolutional channel or a hidden neuron. The \emph{scope} is the set of layers whose units UnAct may edit. We choose it by a rule that refers to the classifier rather than to a layer type: the scope is the late layers whose units write to the representation the classifier reads. These features are the most class-specific, and earlier, shared features stay untouched. The rule can be stated for any architecture with a classifier head; only the number of late layers to include depends on the architecture.

\paragraph{ResNet-18.} The classifier reads the global average of the 512 output channels of the last residual block, so the scope is that block and each channel is a unit. $\mathrm{In}_j$ is the filter producing channel $j$ with its BatchNorm scale and shift, and $\mathrm{Out}_j$ is the classifier's weight column for channel $j$.

\paragraph{ViT-B/16.} The classifier reads the \texttt{[CLS]} token, to which every block writes through a shared residual stream. The units are the MLP hidden neurons of the last four blocks, which work better than the last two (Appendix~\ref{app:vit}). $\mathrm{In}_j$ is the neuron's row of the first MLP projection with its bias, and $\mathrm{Out}_j$ is its column of the second. Residual-stream dimensions are not units, because every block writes to them.

\subsection{Unit Scores}
\label{sec:scoring}

The score $s_j$ estimates how much of the forget images' signal unit $j$ carries to the classifier, that is, how strongly the model uses unit $j$ when it sees class $c$. It needs only forward passes over $\mathcal{D}_f$ and is recomputed on the edited model every round. It does not compare the forget class with the retained ones, so a unit that responds to every class also scores high. In practice, the units above the high percentiles UnAct uses are strongly class-specific (Appendix~\ref{app:selectivity}).

\paragraph{ResNet-18.} The score is the channel's mean activation over the forget images, $s_j = \frac{1}{n}\sum_{x \in \mathcal{D}_f} a_j(x)$, where $a_j(x)$ is the post-ReLU output of channel $j$ for image $x$, averaged over spatial positions, which is exactly the feature the classifier receives.

\paragraph{ViT-B/16.} The same activation score, the signed mean GELU output, leaves the forget class almost intact on ViT-B/16 (Appendix~\ref{app:vit}). GELU explains part of this. Most late-block neurons have a negative mean GELU output, so the selected set contains negative neurons whenever fewer than $(100-p)\%$ are positive; attenuation moves such a neuron toward zero and raises its score, so it is reselected every round. Scoring only the positive part of the GELU output avoids this but still fails ($\Delta = \numVitVariantFfnPosLastfour$, Table~\ref{tab:vit-variants}). The deeper cause is how a neuron reaches the classifier: only through its outgoing weights, which write into the shared residual stream that the final LayerNorm rescales. Its activation therefore does not show how much it contributes to class $c$. We instead score each neuron by its contribution to the class-$c$ logit at \texttt{[CLS]}, $s_j = \frac{1}{n}\sum_{x \in \mathcal{D}_f} g_j(x)\,(\mathbf{w}_j^{\top}\mathbf{u}_c)/\sigma(x)$, where $g_j(x)$ is the neuron's GELU output at \texttt{[CLS]}, $\mathbf{w}_j$ its outgoing weights, $\mathbf{u}_c$ the direction of the class-$c$ logit in the residual stream, and $\sigma(x)$ the final LayerNorm's scale. This is the neuron's direct logit attribution~\citep{elhage2021mathematical}, a first-order attribution that ignores paths through later blocks, and it needs no backward pass (Appendix~\ref{app:vit}). The class $c$ is the one with the highest mean logit on $\mathcal{D}_f$ under the original model. No label is read, but this is effectively a pseudo-label that assumes a single-class forget set (Appendix~\ref{app:multiclass}).

\paragraph{What carries over.} Both scores measure the forget signal a unit sends to the classifier, so the choice depends on how units reach it. We expect the activation score to suit non-negative units the classifier reads directly and the class-token score to suit residual-stream transformers, but have tested one model of each kind.

\subsection{Selection and Iterative Attenuation}
\label{sec:attenuation}

Each round $t$ selects the units whose scores lie at or above the $p$-th percentile of the scope's scores, $\mathcal{M}_t=\{\,j : s_j \geq Q_p(s)\,\}$, which is about the top $(100-p)\%$ (ties: Appendix~\ref{app:proofs}). For every $j\in\mathcal{M}_t$, UnAct multiplies the parameters in $\mathrm{In}_j$ and $\mathrm{Out}_j$ by $\gamma$, with $0<\gamma<1$, weakening the unit without removing it. On ResNet-18 this is not a uniform rescaling of the channel, because BatchNorm's running statistics stay frozen (Appendix~\ref{app:proofs}). At a high percentile one round does not remove the class (Table~\ref{tab:unact-grid}), partly because on ResNet-18 the identity shortcut still carries part of a selected channel's signal (Appendix~\ref{app:proofs}); at lower percentiles one round can suffice, and CIFAR-100's selected configuration uses $k=1$. UnAct therefore repeats scoring and attenuation for $k$ rounds. An attenuated unit's score drops, so later rounds mostly select new units: iteration widens the edit rather than deepening it on a fixed set. By a union bound, the edited fraction of the scope after $k$ rounds is at most $k$ times the per-round fraction, and measured coverage stays close to this bound for small $k$ (Appendix~\ref{app:coverage}). Thus $(p,k)$ controls the breadth of the edit and $\gamma$ its strength. Because selection is by percentile, these bounds hold whatever the number of forget images or the scale of the scores. SSD's threshold rule has no such bound: the number of parameters with $[\mathcal{D}_f]_i > \alpha[\mathcal{D}]_i$ depends on a statistic estimated from the forget images.

\paragraph{Algorithm and cost.}\label{sec:algorithm} Algorithm~\ref{alg:unact} summarizes the procedure. A request costs $k$ forward passes over the $n$ forget images and no backward pass, optimizer state or retained data, so its cost grows with $|\mathcal{D}_f|$ rather than with the training set. Given the model, $\mathcal{D}_f$ and $(p,\gamma,k)$, the procedure is deterministic.

\begin{algorithm}[t]
\caption{\textsc{UnAct}}
\label{alg:unact}
\begin{algorithmic}[1]
\Require model $f_\theta$; forget images $\mathcal{D}_f$; scope; percentile $p$; factor $\gamma$; rounds $k$
\State set $f_\theta$ to inference mode \Comment{BatchNorm statistics frozen, dropout off}
\For{$t = 1$ \textbf{to} $k$}
  \State compute $s_j$ for every unit in the scope \Comment{Section~\ref{sec:scoring}; forward passes only}
  \State $\mathcal{M}_t \gets \{\, j : s_j \geq Q_{p}(s) \,\}$ \Comment{ViT: top units by rank, ties by index}
  \State $\mathrm{In}_j \gets \gamma\,\mathrm{In}_j$, \; $\mathrm{Out}_j \gets \gamma\,\mathrm{Out}_j$ \; for all $j \in \mathcal{M}_t$
\EndFor
\State \Return $f_{\theta'}$
\end{algorithmic}
\end{algorithm}

\section{Results}
\label{sec:results}

\subsection{Experimental setup}
\label{sec:setup}

\paragraph{Models and data.} We train ResNet-18 \citep{he2016deep} with the
$3{\times}3$ CIFAR stem from scratch on CIFAR-10, CIFAR-20 (CIFAR-100 with its 20
superclass labels) and CIFAR-100 \citep{krizhevsky2009learning}, reaching
\numOrigAccCten\%, \numOrigAccCtwenty\% and \numOrigAccChundred\% test accuracy. We also fine-tune ViT-B/16 \citep{dosovitskiy2021image}, pre-trained
on ImageNet-1k, on CIFAR-10 at $224{\times}224$ (Appendix~\ref{app:setup}).
Each request removes one class, and unless stated otherwise $\mathcal{D}_f$
contains all training images from that class. We fix five forget classes per
dataset before any experiment and report averages over them; for each, a model
retrained from scratch on the retain set with the same recipe is the reference.

\paragraph{Metric.} $A_r$ and $A_f$ denote test accuracy on the retained and
forgotten classes, respectively, following SSD and LFSSD
\citep{foster2024fast,foster2024loss}; they are measured on the test split and
are distinct from the training sets $\mathcal{D}_r$ and $\mathcal{D}_f$. Following prior work that reports gaps
relative to a retrained model \citep{jia2023model,fan2024salun,huang2024unified},
we summarize each method using
$\Delta = |A_r - A_r^{\mathrm{gold}}| + |A_f - A_f^{\mathrm{gold}}|$,
measured in percentage points. We compute $\Delta$ separately for each forget
class and then average. $\Delta=0$ only when the edited model matches the
retrained model on both terms, so it penalizes lost retain accuracy as well as
excessive deviation on the forget classes. It is a two-term analogue of the Avg.\ Gap
\citep{fan2024salun} and, like ToW \citep{zhao2024makes}, is also used for
hyperparameter selection. We additionally report an entropy-based membership
inference attack (MIA) following SSD (Appendix~\ref{app:setup}).

\paragraph{Baselines and tuning.} We compare UnAct with two retrain-free
methods closely related to our setting: SSD \citep{foster2024fast} and its
label-free variant LFSSD \citep{foster2024loss}, implemented according to their
respective papers. SSD uses labels, gradients and a parameter importance over
the full training set; LFSSD drops the labels but keeps the rest. On ResNet-18, each method has one
operating point per dataset: the configuration with the lowest mean $\Delta$
over the five forget classes on a grid fixed before evaluation (100
configurations for UnAct and 90 each for SSD and LFSSD, with SSD's selected
$\alpha$ restricted to the range reported in its paper; Appendix~\ref{app:grids}).
\emph{All subsequent experiments reuse these operating points without
re-tuning.} The selected points also transfer to three held-out classes on which
SSD's original paper reports failures (Appendix~\ref{app:e11}). In addition, a
selection rule that does not require a retrained model selects the same operating
point in \numRfSameCount{} of \numRfCells{} method--dataset pairs
(Appendix~\ref{app:selection}). On ViT-B/16, SSD uses the best point from an $\alpha$--$\lambda$ grid, while
UnAct's configuration is selected using one forget class
(Appendix~\ref{app:vit}). We did not run LFSSD on ViT-B/16, whose importance pass over all 50,000 training
images at $224{\times}224$ takes \numVitFsTimeSSDImportance\,s for SSD; SSD's ViT
grid fixes the importance batch at 128. Accuracies are evaluated in fp32 (bfloat16
for the ViT-B/16 relearning curves and first selection stage; Appendices~\ref{app:vit}
and~\ref{app:relearn}) on NVIDIA L4 or A10 GPUs; the two GPUs agree to within one
test image on the same checkpoint (Appendix~\ref{app:setup}). All timing
measurements use one L4 GPU. All experiments use seed 42.

\paragraph{Questions.} We evaluate four progressively harder settings. (Q1)
Given all training images of the forget class, how closely does each method
approach retraining? (Q2) How many forget images does each method require, and
does performance remain stable when only a few are available? (Q3) Do requested
classes remain forgotten when requests are applied sequentially? (Q4) What is
the computational cost of each request? We evaluate these questions on both
architectures.

\subsection{Findings}
\label{sec:findings}\label{sec:data-efficiency}\label{sec:cost}

\begin{table}[t]
\centering
\caption{Removing one class with the full forget set, mean over the five forget
classes. $A_r$, $A_f$: test accuracy on the retained and forgotten classes;
$\Delta$: distance to the retrained model (lower is better; best method in bold).
Each method runs at its selected operating point; n/r: not run. ResNet-18
standard deviations, MIA, Avg.\ Gap and modified-parameter fractions are in
Table~\ref{tab:main-full}; ViT-B/16 details are in Table~\ref{tab:vit}.}
\label{tab:main}
\small
\setlength{\tabcolsep}{3pt}
\begin{tabular}{@{}lcccccccccccc@{}}
\toprule
& \multicolumn{9}{c}{ResNet-18} & \multicolumn{3}{c}{ViT-B/16} \\
\cmidrule(lr){2-10}\cmidrule(l){11-13}
& \multicolumn{3}{c}{CIFAR-10} & \multicolumn{3}{c}{CIFAR-20} & \multicolumn{3}{c}{CIFAR-100} & \multicolumn{3}{c}{CIFAR-10} \\
\cmidrule(lr){2-4}\cmidrule(lr){5-7}\cmidrule(lr){8-10}\cmidrule(l){11-13}
Method & $A_r$ & $A_f$ & $\Delta\!\downarrow$ & $A_r$ & $A_f$ & $\Delta\!\downarrow$ & $A_r$ & $A_f$ & $\Delta\!\downarrow$ & $A_r$ & $A_f$ & $\Delta\!\downarrow$ \\
\midrule
Original model & 95.16 & 93.28 & 93.72 & 85.15 & 88.96 & 89.50 & 77.31 & 81.00 & 81.39 & 98.33 & 97.92 & 98.11 \\
Retrain & 95.60 & 0.00 & --- & 85.69 & 0.00 & --- & 77.71 & 0.00 & --- & 98.46 & 0.00 & --- \\
\midrule
SSD & 95.44 & 0.00 & 0.29 & 85.43 & 0.00 & 0.27 & 77.20 & 0.00 & 0.50 & 78.89 & 14.16 & 33.74 \\
LFSSD & 95.66 & 0.00 & 0.11 & 85.45 & 0.00 & \textbf{0.24} & 77.33 & 0.00 & \textbf{0.38} & n/r & n/r & n/r \\
\textbf{UnAct} (ours) & 95.59 & 0.00 & \textbf{0.08} & 84.90 & 0.00 & 0.80 & 76.56 & 0.00 & 1.15 & 91.93 & 4.96 & \textbf{11.49} \\
\bottomrule
\end{tabular}
\end{table}

\paragraph{(Q1) With the full forget set, UnAct is within about one point of
retraining on ResNet-18 and is clearly closest on ViT-B/16.} On ResNet-18, all three methods drive forget
accuracy to \numDfCtenUnAct\% on every dataset and stay within about one point of
the retrained model (Table~\ref{tab:main}). UnAct is the closest on CIFAR-10
($\Delta = \numDeltaCtenUnAct$, against \numDeltaCtenLFSSD{} for LFSSD and
\numDeltaCtenSSD{} for SSD), and LFSSD on CIFAR-20 and CIFAR-100
($\numDeltaCtwentyLFSSD$ and $\numDeltaChundredLFSSD$, against UnAct's
$\numDeltaCtwentyUnAct$ and $\numDeltaChundredUnAct$); UnAct reaches this regime
without labels, gradients, retained data or a pass over the training set. On
ViT-B/16, where the scope, the score (with a pseudo-label) and the tie-breaking are
adapted (Sections~\ref{sec:scope}--\ref{sec:attenuation}), UnAct is \numVitRatioSSDUnAct{} times closer to
retraining than SSD ($\Delta = \numVitDeltaUnAct$ against $\numVitDeltaSSD$) and
closer on \numVitCloserUnAct{} of the five classes; on the four classes its
configuration was not chosen on, $\Delta$ is \numVitDeltaExclThreeUnAct{} against
\numVitDeltaExclThreeSSD. The gap comes from safety: UnAct's lowest per-class
retain accuracy is \numVitWorstDrUnAct\%, whereas SSD collapses one class to
\numVitWorstDrSSD\% and leaves another largely unforgotten
(Tables~\ref{tab:vit} and~\ref{tab:vit-variants}). The same procedure thus carries over from a
convolutional network to a transformer, although UnAct's mean retain accuracy on
ViT-B/16 remains \numVitDrGapUnAct{} points below the retrained model's. The membership attack does not separate
the methods: every method drives it to at most \numMIAMaxUnlearn\% on ResNet-18,
below the retrained models' rates, so we do not use it to rank them
(Table~\ref{tab:main-full}).

\begin{figure}[t]
\begin{center}
\includegraphics[width=\linewidth]{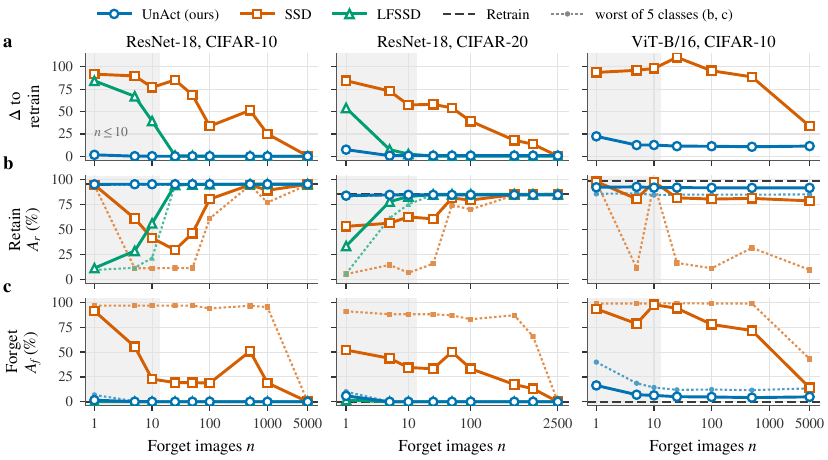}
\end{center}
\caption{\textbf{Forgetting from few images.} Each method runs at its selected
configuration at every forget-set size $n$ (log scale; the last $n$ is the whole
class). (a) $\Delta$, mean over the five forget classes. (b) Retain and (c) forget
accuracy: solid, mean; dotted, the worst of the five classes (lowest $A_r$,
highest $A_f$); dashed, the retrained models. Read (b) and (c) together: a
collapsed network also reaches $A_f = 0$. The shaded band marks $n \leq 10$.
CIFAR-100 and a log-scale $\Delta$ are in Figure~\ref{fig:forget-size-full}.}
\label{fig:forget-size}
\end{figure}

\paragraph{(Q2) With few forget images, UnAct stays safe where the baselines
collapse.} We vary the number $n$ of forget images each method receives, from one
to the whole class, and still evaluate the removal of the whole class
(Figure~\ref{fig:forget-size}). UnAct uses only the $n$ images, whereas SSD and
LFSSD also receive, at every $n$, a parameter importance computed from gradients
over the entire training set, as their papers prescribe, so the comparison gives
the baselines strictly more information than UnAct. Even so, over all $n$, forget classes and
ResNet-18 datasets (\numUnActRunsEthree{} runs), UnAct's retain accuracy is never
more than \numWorstDropUnActMax{} points below the retrained model's. SSD and
LFSSD, at the operating points that serve them well on the full class, damage the
network when given few images: on CIFAR-10 their per-class retain accuracy falls
by up to \numWorstDropSSDCten{} and \numWorstDropLFSSDCten{} points, and their
mean retain accuracy drops to \numMinDrSSDCten\% and \numMinDrLFSSDCten\%, against
UnAct's lowest mean of \numMinDrUnActCten\%. On ViT-B/16, UnAct's largest
per-class drop is \numVitFsWorstDropUnAct{} points against \numVitFsWorstDropSSD{}
for SSD. Row (c) shows the complementary failure: on the classes it does not
collapse, SSD leaves the forget class largely unforgotten until it sees most of
the class, whereas UnAct's forget accuracy on ResNet-18 is near zero from $n=5$.

\paragraph{Five images are enough for UnAct.} With $n=5$, UnAct's $\Delta$ is
already \numDeltaUnActCtenNfive, \numDeltaUnActCtwentyNfive{} and
\numDeltaUnActChundredNfive{} on the three ResNet-18 datasets and
\numVitFsDeltaUnActNfive{} on ViT-B/16, within about one point of its full-class
value there. On CIFAR-10 and CIFAR-20 it is the closest method at every $n \leq
10$; at $n = 5$ it is \numRatioLFSSDUnActCtenNfive{} and
\numRatioLFSSDUnActCtwentyNfive{} times closer than LFSSD, and
\numRatioSSDUnActCtenNfive{} and \numRatioSSDUnActCtwentyNfive{} times closer than
SSD. SSD needs the entire class to come within a factor of two of its best
$\Delta$ on ResNet-18 (Appendix~\ref{app:data-efficiency}), and on ViT-B/16 its
$\Delta$ is at least \numVitFsDeltaSSDNfivehundred{} at every $n \leq 500$ (Figure~\ref{fig:vit-forget-size}). LFSSD,
drawing on its full-training-set gradients, reaches a lower $\Delta$ than UnAct on
CIFAR-100 from $n=5$ ($\numDeltaLFSSDChundredNfive$ against
$\numDeltaUnActChundredNfive$), on CIFAR-20 from $n=25$ and on CIFAR-10 at $n=100$
(0.09 against 0.17), but with few images it
collapses some classes on CIFAR-10 and CIFAR-20, as SSD does. Re-selecting SSD's $\alpha$ at every $n$ within
the range its paper reports, with the retrained model as an oracle, still leaves
its $\Delta$ at $n=5$ at \numOracleSSDNfiveCten, \numOracleSSDNfiveCtwenty{} and
\numOracleSSDNfiveChundred{}, far above UnAct's untuned values
(Table~\ref{tab:ssd-oracle}).

\paragraph{(Q3) Sequential requests.} Applied five times in sequence to the
already-edited ResNet-18, UnAct keeps every requested class forgotten on CIFAR-10
and CIFAR-20 (forget accuracy at most \numSeqDfMaxUnActCtenCtwenty\% over five
class orders), whereas SSD's forget accuracy on the requested classes exceeds
50\% by the fifth request. UnAct's retain accuracy erodes on CIFAR-20 after the
third request, to \numSeqDrUnActCtwentyRfive\% after the fifth against
\numSeqDrSSDCtwentyRfive\% for SSD, and on CIFAR-100 SSD is the better choice
(Appendix~\ref{app:multiclass}).

\paragraph{(Q4) Cost.} UnAct's cost is $k$ forward passes over the forget images;
it never reads the training set and stores nothing between requests. On
ResNet-18, the best one-round configuration is within
\numKoneMinusSelectedCten{}, \numKoneMinusSelectedCtwenty{} and
\numKoneMinusSelectedChundred{} of the selected $\Delta$ and removes a whole class
in \numTimeUnActCtenKone{}, \numTimeUnActCtwentyKone{} and
\numTimeUnActChundredKone{}\,s: \numSpeedupAmortCtenKone{},
\numSpeedupAmortCtwentyKone{} and \numSpeedupAmortChundredKone{} times faster than
SSD with its importance precomputed, and
\numSpeedupColdCtenKone{}--\numSpeedupColdChundredKone{} times faster from cold
(Appendix~\ref{app:cost}). On ViT-B/16 a
full-class request takes UnAct \numVitFsTimeUnActFull\,s against SSD's
\numVitFsTimeSSDColdFull\,s, of which \numVitFsTimeSSDImportance\,s is the
importance pass (\numVitFsSpeedupCold$\times$ faster from cold,
\numVitFsSpeedupAmort$\times$ with the importance precomputed), and five images
take \numVitFsTimeUnActNfive\,s. LFSSD has the same two-pass structure as SSD.

\paragraph{What $\Delta$ certifies.} A trivial baseline that zeroes the class
logit reaches $\Delta = \numDepthDeltaMaskCten$, \numDepthDeltaMaskCtwenty{} and
\numDepthDeltaMaskChundred{} on ResNet-18, as close to retraining as any method,
and a linear probe recovers the forget class from the penultimate features with
AUROC of at least \numDepthProbeMin{} after every method, including ours
(Table~\ref{tab:depth}). $\Delta$ therefore certifies that a class is no longer
predicted, not that it has been removed, and we use it only in that sense.

\section{Discussion and conclusion}
\label{sec:discussion}

We asked how little information a retrain-free method needs to remove a class
from a trained classifier. UnAct answers with a forward-only edit: it scores
late-layer units by the forget signal they send to the classifier, selects a fixed
fraction of them by rank, attenuates them softly and rescores the edited model,
reading no gradients, labels or retained data.

\paragraph{Robustness when forget data is scarce.} With the whole class, UnAct
is competitive with SSD and LFSSD on ResNet-18 while using strictly less information.
The forget-set-size experiments are where the methods separate. There the baselines
receive more information, an importance over the full training set at every $n$,
but keep their full-class operating points; only SSD is also re-tuned per $n$. Even so, on ResNet-18 UnAct is the only method that is safe
at every forget-set size, and with at most ten images it is the closest method to
retraining on CIFAR-10 and CIFAR-20, a gap that oracle re-selection of SSD's
threshold does not close. As $n$ grows, LFSSD becomes comparable to UnAct and is
closer on CIFAR-100 from $n=5$, on CIFAR-20 from $n=25$ and on CIFAR-10 at $n=100$, while SSD needs the entire class to approach its best value.
On ViT-B/16, UnAct is closer to retraining than SSD at every forget-set size.

\paragraph{Limitations.} UnAct suppresses the forget class rather than erasing
it: a few fine-tuning steps on a handful of forget images restore much of the
class, as they do for SSD. It is designed for one class per request, since its
scoring presumes that the forget images come from a single class. With the
whole class available, it is slightly less close to retraining than SSD and
LFSSD on CIFAR-20 and CIFAR-100 and edits more parameters than either, and on
ViT-B/16 its retain accuracy remains below that of the retrained model. Finally, the score must be adapted to how units reach the
classifier (Section~\ref{sec:scoring}). Our evidence covers CIFAR-scale data,
two architectures and one training seed, with operating points selected on the
test split for all methods.

\subsection*{AI use statement}

In this work, we used generative AI tools to debug code, to format LaTeX tables and the algorithm, to identify related literature, and to review and edit the manuscript text, including the proper wording and framing of results the authors had already obtained. We have not used generative AI tools to propose the method, formulate hypotheses, or generate or clean datasets, and synthetic data generation, translation, and qualitative data analysis are not applicable to this work. We have reviewed all AI-assisted work: AI-suggested code fixes were reviewed and tested by the authors, the baseline was cross-checked manually against the reference implementation, all results come from code written and executed by the authors, and every edited claim and added citation was checked by the authors against the results files and the cited papers. We take responsibility for the final content of this work, including text, claims or artifacts produced with the aid of generative AI.



\subsection*{Reproducibility statement}

Section~\ref{sec:method} and Algorithm~\ref{alg:unact} describe UnAct, and
Appendix~\ref{app:proofs} gives further implementation details.
Section~\ref{sec:setup} and Appendix~\ref{app:setup} describe the datasets, forget
classes, training setups, evaluation protocol, hardware and baseline
implementations. Appendix~\ref{app:grids} lists the hyperparameter grids and the
selected operating points. Source code, experiment scripts and results
files are provided in the supplementary material and at the link in the abstract.


\bibliography{iclr2027_conference}
\bibliographystyle{iclr2027_conference}

\appendix
\section*{Appendix}
\addcontentsline{toc}{section}{Appendix}

\section{Method details}
\label{app:proofs}
\subsection{The identity path in residual blocks}

In ResNet-18 the score is measured on the block output, which includes the
identity shortcut, while the update edits only the residual branch. A selected
channel therefore still receives signal through the shortcut, and one round
does not suppress it completely. This is one reason several rounds help on
ResNet-18. Fully connected layers have no such path.

\subsection{What attenuation does on ResNet-18}

$\mathrm{In}_j$ is the channel's \texttt{conv2} filter together with its
BatchNorm scale $w$ and shift $b$, and all three are multiplied by $\gamma$. The
running statistics $\mu$ and $\sigma$ stay frozen, so the channel's pre-activation
$w(z-\mu)/\sigma+b$ becomes $\gamma[w(\gamma z-\mu)/\sigma+b]$, not $\gamma$
times its original value. The classifier column $\mathrm{Out}_j$ is scaled
exactly, and the classifier bias is not edited.

\subsection{Selection count and ties}

With $Q_p$ the linearly interpolated percentile, the selection rule of
Section~\ref{sec:attenuation} selects $N - \lceil (N-1)\,p/100 \rceil$ of $N$ units when the scores are
distinct, which is slightly more than $N(1-p/100)$. When many scores are tied,
the rule selects every tied unit. On ViT, many scores are exactly zero at the
class token, so for the class-token score (Section~\ref{sec:scoring}) we select the same number of units
by rank, breaking ties by index.

\section{Training and evaluation details}
\label{app:setup}

\paragraph{ResNet-18.} We use the CIFAR stem, a $3{\times}3$ stride-1 first
convolution with no max-pool; the stock ImageNet stem would downsample
$32{\times}32$ inputs to $8{\times}8$ before the first residual block. We train
for 100 epochs with SGD, momentum 0.9, weight decay $5\times10^{-4}$, batch size
256, and a cosine schedule peaking at a learning rate of 0.2 after 5 linear
warm-up epochs, with random crops (padding 4) and horizontal flips. Training uses
bfloat16 autocast; every reported accuracy is computed in fp32. Each retrained
model uses the same recipe on the retain set alone.

\paragraph{ViT-B/16.} We fine-tune torchvision's \texttt{vit\_b\_16} with
\texttt{IMAGENET1K\_V1} weights, pre-trained on ImageNet-1k only. The SSD paper
uses a ViT-B/16 pre-trained on ImageNet-21k, so our ViT numbers are not a
reproduction of its ViT results. Images are upsampled from $32{\times}32$ to
$224{\times}224$ after the same crop and flip augmentation as for ResNet-18. We
train for 8 epochs with SGD, momentum 0.9, no weight decay, learning rate 0.01 at
batch size 128, gradient clipping at norm 1.0, and a linear warm-up over 6\% of
the steps followed by cosine decay, under bfloat16 autocast. The five retrained
ViT models use the same recipe on the retain set. ViT training is not
bit-reproducible, because the attention backward pass has no deterministic
kernel; all unlearning and evaluation passes are forward-only and deterministic.

\paragraph{Label-free check.} For forget class 3 of each dataset we run each
method twice, once with the true forget-set labels and once with every label
replaced by a constant, and compare the edited weights. UnAct runs at its
selected configurations; SSD and LFSSD at $\alpha=10$, $\lambda=1$ and importance
batch 256. UnAct's and LFSSD's weights are bit-identical in every case, and SSD's
differ in \numLabelFreeSSDDiffMinM{}--\numLabelFreeSSDDiffMaxM{} million
parameters.

\paragraph{Forget classes and evaluation.} The forget classes, fixed before any
experiment, are $\{0,2,3,5,8\}$ for CIFAR-10 (also used for ViT-B/16),
$\{3,4,10,14,19\}$ for CIFAR-20 and $\{3,20,51,69,85\}$ for CIFAR-100.
Evaluation uses fp32, deterministic kernels and a fixed batch size. Rows were
evaluated on an NVIDIA L4 or A10; on the same checkpoint the two devices differ
by at most one test image (about 0.01 points of $A_r$, and 0.1, 0.2 and 1.0
points of $A_f$ on CIFAR-10, CIFAR-20 and CIFAR-100), and the tables that mix
devices say so. The edit itself is recomputed on the GPU that runs each
experiment, so full-class $\Delta$ values taken from the A10 runs
(Tables~\ref{tab:ssd-grid}, \ref{tab:unact-grid}, \ref{tab:selection},
\ref{tab:depth}, \ref{tab:data-efficiency}, \ref{tab:ssd-oracle} and
\ref{tab:cost}) can differ from the L4 run of Table~\ref{tab:main} by 0.01, for
example 0.51 against 0.50 for SSD and 1.14 against 1.15 for UnAct on CIFAR-100.

\paragraph{Membership inference.} As in SSD, a logistic regression on
prediction entropy is trained to separate retain-set training images from test
images; MIA is the fraction of forget-set training images it labels as members.

\paragraph{Baselines.} SSD and LFSSD compute their importance over the full
training set once per model and reuse it for every forget-set size. Their grids
include the importance batch size because the importance is averaged over
batches, so the meaning of $\alpha$ depends on it. Our LFSSD
takes the absolute value of the batch-mean gradient of $\lVert f(x)\rVert_2^2$,
as the authors' reference code does, rather than averaging per-sample absolute
gradients as in the LFSSD paper's equation; within-batch cancellation therefore
makes its importance depend on the batch size, which is one axis of its grid. On
ViT-B/16, SSD's grid is $\alpha \in \{2,5,6,7,8,9,10,15,25,50\}$ and $\lambda \in
\{0.1,0.5,1\}$ at importance batch 128, selected on all five classes with
bfloat16 evaluation on 2{,}000 retain test images and re-evaluated at fp32 on the
full split.

\section{Hyperparameter grids and sensitivity}
\label{app:grids}

\begin{table}[t]
\centering
\caption{Complete SSD grid, $\Delta$ averaged over the five
forget classes, at each dataset's selected $\lambda$ (rows: $\alpha$, columns:
importance batch size). The selected operating point is in bold; the full
$\lambda$ axis is in the results file. $\alpha<5$ lies outside the range SSD's
paper reports and was never selected.}
\label{tab:ssd-grid}
\scriptsize
\begin{minipage}{0.32\textwidth}\centering
CIFAR-10 ($\lambda=1$)\\[2pt]
\begin{tabular}{rccc}
\toprule
$\alpha$ & 64 & 256 & 512 \\
\midrule
2 & 32.75 & 60.06 & 75.43 \\
4 & 18.41 & 20.66 & 28.09 \\
6 & 2.99 & 8.53 & 11.87 \\
8 & 31.23 & \textbf{0.29} & 0.61 \\
10 & 90.96 & 93.49 & 93.36 \\
12 & 93.72 & 93.73 & 93.74 \\
15 & 93.74 & 93.70 & 93.74 \\
20 & 93.74 & 93.71 & 93.72 \\
30 & 93.72 & 93.72 & 93.72 \\
50 & 93.72 & 93.72 & 93.72 \\
\bottomrule
\end{tabular}
\end{minipage}\hfill
\begin{minipage}{0.32\textwidth}\centering
CIFAR-20 ($\lambda=1$)\\[2pt]
\begin{tabular}{rccc}
\toprule
$\alpha$ & 64 & 256 & 512 \\
\midrule
2 & 3.59 & 12.05 & 26.45 \\
4 & 1.02 & 3.62 & 7.58 \\
6 & 0.54 & 1.64 & 3.08 \\
8 & 0.34 & 0.83 & 1.67 \\
10 & 4.76 & 0.45 & 0.91 \\
12 & 11.98 & 0.32 & 0.44 \\
15 & 30.12 & 5.16 & \textbf{0.27} \\
20 & 89.30 & 88.79 & 89.68 \\
30 & 89.50 & 89.50 & 89.50 \\
50 & 89.50 & 89.50 & 89.50 \\
\bottomrule
\end{tabular}
\end{minipage}\hfill
\begin{minipage}{0.32\textwidth}\centering
CIFAR-100 ($\lambda=0.5$)\\[2pt]
\begin{tabular}{rccc}
\toprule
$\alpha$ & 64 & 256 & 512 \\
\midrule
2 & 50.46 & 75.42 & 76.10 \\
4 & 23.09 & 67.60 & 74.74 \\
6 & 13.01 & 53.62 & 71.85 \\
8 & 8.72 & 33.43 & 63.89 \\
10 & 5.89 & 24.32 & 54.21 \\
12 & 4.21 & 18.74 & 42.15 \\
15 & 2.29 & 11.16 & 29.62 \\
20 & 1.40 & 8.09 & 17.90 \\
30 & 0.86 & 2.50 & 7.43 \\
50 & \textbf{0.51} & 0.75 & 1.74 \\
\bottomrule
\end{tabular}
\end{minipage}\hfill
\end{table}

\begin{table}[t]
\centering
\caption{Complete LFSSD grid, $\Delta$ averaged over the five
forget classes, at each dataset's selected $\lambda$ (rows: $\alpha$, columns:
importance batch size). The selected operating point is in bold; the full
$\lambda$ axis is in the results file. $\alpha<5$ lies outside the range SSD's
paper reports and was never selected.}
\label{tab:lfssd-grid}
\scriptsize
\begin{minipage}{0.32\textwidth}\centering
CIFAR-10 ($\lambda=0.1$)\\[2pt]
\begin{tabular}{rccc}
\toprule
$\alpha$ & 64 & 256 & 512 \\
\midrule
2 & 83.98 & 84.22 & 84.49 \\
4 & 10.54 & 54.27 & 74.67 \\
6 & \textbf{0.11} & 0.78 & 6.54 \\
8 & 3.84 & 8.29 & 7.31 \\
10 & 93.72 & 93.72 & 93.72 \\
12 & 93.72 & 93.72 & 93.72 \\
15 & 93.72 & 93.72 & 93.72 \\
20 & 93.72 & 93.72 & 93.72 \\
30 & 93.72 & 93.72 & 93.72 \\
50 & 93.72 & 93.72 & 93.72 \\
\bottomrule
\end{tabular}
\end{minipage}\hfill
\begin{minipage}{0.32\textwidth}\centering
CIFAR-20 ($\lambda=0.1$)\\[2pt]
\begin{tabular}{rccc}
\toprule
$\alpha$ & 64 & 256 & 512 \\
\midrule
2 & 76.19 & 80.00 & 80.05 \\
4 & 11.48 & 51.54 & 69.57 \\
6 & 1.18 & 10.87 & 22.37 \\
8 & 0.45 & 1.33 & 4.69 \\
10 & \textbf{0.24} & 0.41 & 0.85 \\
12 & 0.31 & 0.31 & 0.36 \\
15 & 3.90 & 5.70 & 4.57 \\
20 & 89.49 & 89.49 & 89.49 \\
30 & 89.49 & 89.49 & 89.49 \\
50 & 89.49 & 89.49 & 89.49 \\
\bottomrule
\end{tabular}
\end{minipage}\hfill
\begin{minipage}{0.32\textwidth}\centering
CIFAR-100 ($\lambda=0.5$)\\[2pt]
\begin{tabular}{rccc}
\toprule
$\alpha$ & 64 & 256 & 512 \\
\midrule
2 & 76.21 & 76.66 & 76.69 \\
4 & 48.37 & 74.96 & 76.33 \\
6 & 20.58 & 51.70 & 69.22 \\
8 & 3.61 & 27.89 & 43.92 \\
10 & 1.77 & 13.10 & 26.73 \\
12 & 1.24 & 6.22 & 16.03 \\
15 & 0.86 & 1.71 & 5.56 \\
20 & 0.56 & 0.84 & 1.14 \\
30 & \textbf{0.38} & 0.50 & 0.48 \\
50 & 7.77 & 37.15 & 32.77 \\
\bottomrule
\end{tabular}
\end{minipage}\hfill
\end{table}

\begin{table}[t]
\centering
\caption{Complete UnAct grid, $\Delta$ averaged over the five forget
classes, at each dataset's selected $\gamma$ (rows: $p$, columns: rounds $k$).
Selected operating point in bold. The usable region runs diagonally in $(p,k)$:
raising $p$ and raising $k$ trade off at roughly constant coverage
(Eq.~\ref{eq:coverage}). The full $\gamma$ axis is in the results file.}
\label{tab:unact-grid}
\scriptsize
\begin{minipage}{0.32\textwidth}\centering
CIFAR-10 ($\gamma=0.01$)\\[2pt]
\begin{tabular}{rcccc}
\toprule
$p$ & \multicolumn{4}{c}{rounds $k$} \\
 & 1 & 5 & 10 & 20 \\
\midrule
50 & 0.58 & 5.90 & 7.10 & 7.22 \\
75 & 0.15 & 0.36 & 0.48 & 0.46 \\
90 & 61.56 & 0.12 & 0.15 & 0.15 \\
95 & 80.79 & 0.14 & 0.12 & 0.11 \\
99 & 89.96 & 78.66 & 52.65 & \textbf{0.08} \\
\bottomrule
\end{tabular}
\end{minipage}\hfill
\begin{minipage}{0.32\textwidth}\centering
CIFAR-20 ($\gamma=0.1$)\\[2pt]
\begin{tabular}{rcccc}
\toprule
$p$ & \multicolumn{4}{c}{rounds $k$} \\
 & 1 & 5 & 10 & 20 \\
\midrule
50 & 3.47 & 27.78 & 74.17 & 77.69 \\
75 & 0.84 & 5.40 & 13.04 & 20.75 \\
90 & 25.52 & 1.84 & 3.59 & 5.68 \\
95 & 60.71 & 0.82 & 1.64 & 3.00 \\
99 & 82.38 & 55.14 & 18.33 & \textbf{0.80} \\
\bottomrule
\end{tabular}
\end{minipage}\hfill
\begin{minipage}{0.32\textwidth}\centering
CIFAR-100 ($\gamma=0.3$)\\[2pt]
\begin{tabular}{rcccc}
\toprule
$p$ & \multicolumn{4}{c}{rounds $k$} \\
 & 1 & 5 & 10 & 20 \\
\midrule
50 & 17.15 & 60.81 & 73.59 & 75.04 \\
75 & 4.17 & 33.87 & 57.07 & 66.56 \\
90 & \textbf{1.14} & 9.54 & 24.42 & 47.96 \\
95 & 11.62 & 3.21 & 9.49 & 23.83 \\
99 & 58.57 & 6.00 & 1.32 & 2.85 \\
\bottomrule
\end{tabular}
\end{minipage}\hfill
\end{table}

\begin{figure}[t]
\begin{center}
\includegraphics[width=\linewidth]{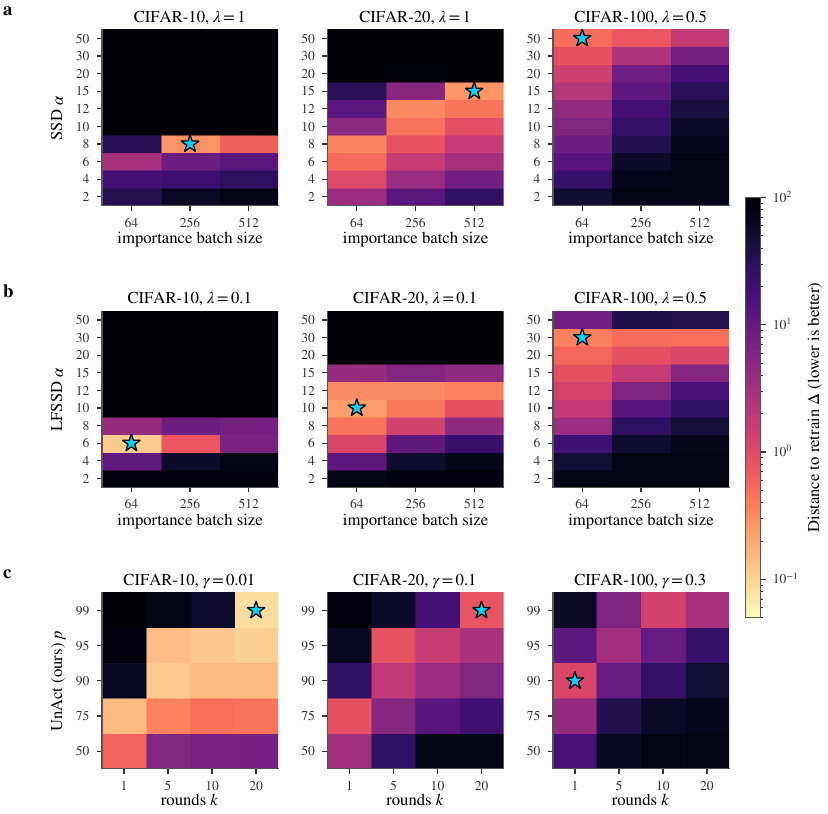}
\end{center}
\caption{Complete search grids, $\Delta$ averaged over the five forget classes,
with stars on the selected configurations. (a) SSD and (b) LFSSD at their
selected $\lambda$; (c) UnAct at its selected $\gamma$. UnAct's usable region
runs diagonally, as Eq.~\eqref{eq:coverage} predicts: raising $p$ and raising $k$
trade off against one another at fixed coverage.}
\label{fig:grids}
\end{figure}

Tables~\ref{tab:ssd-grid}--\ref{tab:unact-grid} and Figure~\ref{fig:grids} report
each method's full grid. Every method needs per-dataset tuning: SSD's selected
$\alpha$ moves from 8 to 15 to 50 across the three datasets, LFSSD's from 6 to 10
to 30, and UnAct's $(p, \gamma, k)$ from $(99, 0.01, 20)$ and $(99, 0.1, 20)$ to
$(90, 0.3, 1)$.

\begin{figure}[t]
\begin{center}
\includegraphics[width=\linewidth]{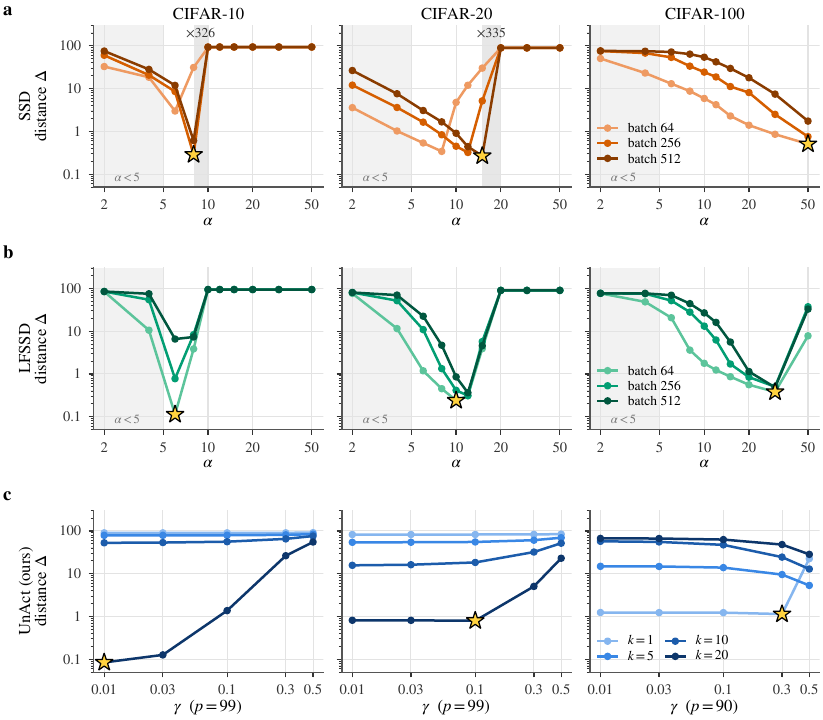}
\end{center}
\caption{One-dimensional sweeps along each method's own axes, $\Delta$ averaged
over the five forget classes, log scale; stars mark the selected configurations.
(a) SSD and (b) LFSSD against $\alpha$, one line per importance batch size; the
shaded region $\alpha<5$ lies outside the range the SSD paper reports. (c) UnAct
against the attenuation factor $\gamma$ at its selected $p$, one line per number
of rounds $k$: flat below a threshold and degrading above it.}
\label{fig:sensitivity}
\end{figure}

Figure~\ref{fig:sensitivity} sweeps each method along its own axes. UnAct is
insensitive to $\gamma$ below a threshold, and its $(p, k)$ trade off through the
coverage $\Gamma_k$ of Eq.~\eqref{eq:coverage} (Appendix~\ref{app:coverage}).

\section{Selection without a retrained model}
\label{app:selection}

\begin{table}[t]
\centering
\caption{Choosing the operating point without a retrained model. $\Delta$ (mean
over the five forget classes) of the configuration each rule picks from the
method's full grid. For class removal the retrained model's forget accuracy is 0
by construction, so the retrain-free rule keeps the configurations with mean forget
accuracy 0 and picks the one with the highest retain accuracy; it uses a labelled
retain evaluation set (here the test split) but no retrained model. Transfer:
the configuration with the lowest mean $\Delta$ on the other two datasets. One
configuration: the lowest mean $\Delta$ over all three datasets.}
\label{tab:selection}
\small
\begin{tabular}{@{}llccc@{}}
\toprule
Method & Selection rule & CIFAR-10 & CIFAR-20 & CIFAR-100 \\
\midrule
SSD & Oracle ($\min\Delta$, needs retraining) & 0.29 & 0.27 & 0.51 \\
 & Retrain-free ($A_f{=}0$, $\max A_r$) & 0.29 & 0.27 & 0.51 \\
 & Transfer from the other two datasets & 31.23 & 0.54 & 32.42 \\
 & One configuration for all three & 2.99 & 0.54 & 12.29 \\
\midrule
LFSSD & Oracle ($\min\Delta$, needs retraining) & 0.11 & 0.24 & 0.38 \\
 & Retrain-free ($A_f{=}0$, $\max A_r$) & 0.11 & 0.24 & 0.38 \\
 & Transfer from the other two datasets & 93.72 & 0.45 & 20.58 \\
 & One configuration for all three & 3.84 & 0.45 & 3.81 \\
\midrule
UnAct & Oracle ($\min\Delta$, needs retraining) & 0.08 & 0.80 & 1.14 \\
 & Retrain-free ($A_f{=}0$, $\max A_r$) & 0.11 & 0.80 & 1.14 \\
 & Transfer from the other two datasets & 1.37 & 0.82 & 3.46 \\
 & One configuration for all three & 0.08 & 0.82 & 3.46 \\
\bottomrule
\end{tabular}
\end{table}

Every operating point in the paper is chosen by $\Delta$, which needs a model
retrained without the forget class. Table~\ref{tab:selection} asks how much this
matters. For class removal the retrained model never predicts the removed class,
so its forget accuracy is 0 in every run we trained, and
$\Delta = |A_r - A_r^{\mathrm{gold}}| + A_f$. The only term that needs retraining
is $A_r^{\mathrm{gold}}$. The retrain-free rule therefore requires forget
accuracy 0 and maximises retain accuracy; it can differ from $\Delta$-selection
when a configuration overshoots the retrained model's retain accuracy, or when the
$\Delta$-optimal configuration leaves a small nonzero forget accuracy. It picks
the $\Delta$-optimal configuration in \numRfSameCount{} of \numRfCells{} cases;
the exception is UnAct on CIFAR-10, where its choice is only \numRfMaxGap{}
worse. The rule still reads a labelled retain evaluation set, here the same test
split we report. A practitioner would use a held-out validation split, so this
shows that retraining is unnecessary, not that the selection is held-out.

Transfer across datasets separates the methods more clearly. A configuration
chosen on the other two datasets, or one configuration shared by all three,
keeps UnAct within $\Delta \leq \numLodoMaxUnAct$ on every dataset for a single
request (the shared configuration is \numSingleCfgUnAct). SSD's and LFSSD's
transferred configurations fail on some dataset ($\Delta$ up to
\numLodoMaxSSD{} and \numLodoMaxLFSSD{}), because their selected $\alpha$ grows
five-fold (LFSSD) and more than six-fold (SSD) from CIFAR-10 to CIFAR-100: an importance ratio does not have
the same meaning on every model, whereas a percentile and an attenuation factor
do.

\section{Held-out classes}
\label{app:e11}

\begin{table}[t]
\centering
\caption{Held-out classes on which SSD's own paper reports failures. Each method
runs at the configuration it selected on the five main forget classes
(transfer, the headline), with $A_r/A_f$ and $\Delta$; ``oracle'' is its best
$\Delta$ over its whole grid on that class. These classes were chosen because a
paper reports SSD failing on them, so they are kept out of every five-class mean.
At SSD's published default ($\alpha=10$, $\lambda=1$) the failure reproduces at
some importance batch sizes: baby, batch 64: $A_r$ 72.8; baby, batch 256: $A_r$ 6.1; baby, batch 512: $A_r$ 2.0; lamp, batch 64: $A_r$ 73.8; lamp, batch 256: $A_r$ 54.9; lamp, batch 512: $A_r$ 34.2; electrical dev., batch 64: $A_r$ 80.3; electrical dev., batch 256: $A_r$ 85.7; electrical dev., batch 512: $A_r$ 85.6.}
\label{tab:e11}
\scriptsize
\setlength{\tabcolsep}{2.8pt}
\begin{tabular}{@{}lcccccccccc@{}}
\toprule
& Retrain & \multicolumn{3}{c}{SSD} & \multicolumn{3}{c}{LFSSD} & \multicolumn{3}{c}{UnAct (ours)} \\
\cmidrule(lr){3-5}\cmidrule(lr){6-8}\cmidrule(l){9-11}
Class & $A_r$ & $A_r/A_f$ & $\Delta$ & oracle & $A_r/A_f$ & $\Delta$ & oracle & $A_r/A_f$ & $\Delta$ & oracle \\
\midrule
CIFAR-100 baby & 77.79 & 77.28/0.00 & 0.51 & 0.51 & 77.39/0.00 & \textbf{0.39} & 0.22 & 76.54/0.00 & 1.25 & 0.74 \\
CIFAR-100 lamp & 77.37 & 77.32/0.00 & 0.05 & 0.04 & 77.45/0.00 & 0.08 & 0.01 & 77.33/0.00 & \textbf{0.04} & 0.04 \\
CIFAR-20 electrical dev. & 86.18 & 85.80/0.00 & \textbf{0.38} & 0.29 & 85.78/0.00 & 0.40 & 0.35 & 85.19/0.20 & 1.19 & 0.98 \\
\bottomrule
\end{tabular}
\end{table}

The SSD paper reports full-class failures on CIFAR-100 \emph{baby} and
\emph{lamp} and on the CIFAR-20 superclass \emph{household electrical devices}.
We keep these classes out of every five-class mean, although two of the randomly
drawn CIFAR-20 class orders of Appendix~\ref{app:multiclass} include
\emph{household electrical devices}, and run each method at the
configuration it selected on the five main classes (Table~\ref{tab:e11}). All
three methods transfer: each forgets the three classes and stays close to
retraining, which indicates that the baselines are not weakened by our
selection. At SSD's published default ($\alpha=10$, $\lambda=1$) the reported
failure reproduces at some importance batch sizes.

\section{Full-class results}
\label{app:main-full}

\begin{table}[t]
\centering
\caption{Full version of Table~\ref{tab:main}: mean $\pm$ s.d.\ over the five
forget classes. Avg.\ Gap follows \citet{fan2024salun} over the three terms we
record, $|\Delta A_f|$, $|\Delta A_r|$ and $|\Delta\mathrm{MIA}|$ with MIA in
points; it omits their training-set retain accuracy. Because every unlearning
method drives MIA to about 0 against a retrained model at 7--30, the MIA term
dominates Avg.\ Gap. ``Params'' is the percentage of parameters modified. LFSSD
rows on CIFAR-20 and CIFAR-100 were evaluated on an A10 rather than the L4; the
two devices differ by at most one test image on the same checkpoint.}
\label{tab:main-full}
\scriptsize
\setlength{\tabcolsep}{2.6pt}
\resizebox{\textwidth}{!}{%
\begin{tabular}{@{}llcccccccc@{}}
\toprule
 & Method & Configuration & $A_r$ & $A_f$ & MIA & $\Delta\downarrow$ & Avg.\ Gap$\downarrow$ & Params (\%) & Eval \\
\midrule
\multirow{5}{*}{CIFAR-10}  & Original model & \texttt{\scriptsize ---} & $95.16${\scriptsize$\,\pm\,0.37$} & $93.28${\scriptsize$\,\pm\,3.32$} & $90.4${\scriptsize$\,\pm\,2.2$} & 93.72{\scriptsize$\,\pm\,3.10$} & 51.25 & --- & L4 \\
  & Retrain & \texttt{\scriptsize 100 epochs} & $95.60${\scriptsize$\,\pm\,0.60$} & $0.00${\scriptsize$\,\pm\,0.00$} & $30.4${\scriptsize$\,\pm\,10.1$} & --- & --- & --- & L4 \\
  & SSD & \texttt{\scriptsize a8.0, l1.0, b256} & $95.44${\scriptsize$\,\pm\,0.38$} & $0.00${\scriptsize$\,\pm\,0.00$} & $0.9${\scriptsize$\,\pm\,1.8$} & 0.29{\scriptsize$\,\pm\,0.47$} & 9.94 & 3.63 & L4 \\
  & LFSSD & \texttt{\scriptsize a6.0, l0.1, b64} & $95.66${\scriptsize$\,\pm\,0.52$} & $0.00${\scriptsize$\,\pm\,0.00$} & $0.0${\scriptsize$\,\pm\,0.1$} & 0.11{\scriptsize$\,\pm\,0.06$} & 10.17 & 3.09 & L4 \\
  & UnAct (ours) & \texttt{\scriptsize p99.0, g0.01, k20} & $95.59${\scriptsize$\,\pm\,0.56$} & $0.00${\scriptsize$\,\pm\,0.00$} & $0.0${\scriptsize$\,\pm\,0.0$} & 0.08{\scriptsize$\,\pm\,0.07$} & 10.17 & 4.80 & L4 \\
\midrule
\multirow{5}{*}{CIFAR-20}  & Original model & \texttt{\scriptsize ---} & $85.15${\scriptsize$\,\pm\,0.17$} & $88.96${\scriptsize$\,\pm\,3.19$} & $94.6${\scriptsize$\,\pm\,1.2$} & 89.50{\scriptsize$\,\pm\,3.08$} & 55.64 & --- & L4 \\
  & Retrain & \texttt{\scriptsize 100 epochs} & $85.69${\scriptsize$\,\pm\,0.38$} & $0.00${\scriptsize$\,\pm\,0.00$} & $17.1${\scriptsize$\,\pm\,8.1$} & --- & --- & --- & L4 \\
  & SSD & \texttt{\scriptsize a15.0, l1.0, b512} & $85.43${\scriptsize$\,\pm\,0.25$} & $0.00${\scriptsize$\,\pm\,0.00$} & $0.2${\scriptsize$\,\pm\,0.3$} & 0.27{\scriptsize$\,\pm\,0.22$} & 5.72 & 0.60 & L4 \\
  & LFSSD & \texttt{\scriptsize a10.0, l0.1, b64} & $85.45${\scriptsize$\,\pm\,0.28$} & $0.00${\scriptsize$\,\pm\,0.00$} & $0.2${\scriptsize$\,\pm\,0.2$} & 0.24{\scriptsize$\,\pm\,0.19$} & 5.74 & 0.79 & A10 \\
  & UnAct (ours) & \texttt{\scriptsize p99.0, g0.1, k20} & $84.90${\scriptsize$\,\pm\,0.22$} & $0.00${\scriptsize$\,\pm\,0.00$} & $0.0${\scriptsize$\,\pm\,0.0$} & 0.80{\scriptsize$\,\pm\,0.38$} & 5.98 & 4.95 & L4 \\
\midrule
\multirow{5}{*}{CIFAR-100}  & Original model & \texttt{\scriptsize ---} & $77.31${\scriptsize$\,\pm\,0.10$} & $81.00${\scriptsize$\,\pm\,10.00$} & $94.4${\scriptsize$\,\pm\,2.6$} & 81.39{\scriptsize$\,\pm\,10.04$} & 56.18 & --- & L4 \\
  & Retrain & \texttt{\scriptsize 100 epochs} & $77.71${\scriptsize$\,\pm\,0.19$} & $0.00${\scriptsize$\,\pm\,0.00$} & $7.2${\scriptsize$\,\pm\,3.8$} & --- & --- & --- & L4 \\
  & SSD & \texttt{\scriptsize a50.0, l0.5, b64} & $77.20${\scriptsize$\,\pm\,0.15$} & $0.00${\scriptsize$\,\pm\,0.00$} & $0.0${\scriptsize$\,\pm\,0.0$} & 0.50{\scriptsize$\,\pm\,0.20$} & 2.57 & 0.23 & L4 \\
  & LFSSD & \texttt{\scriptsize a30.0, l0.5, b64} & $77.33${\scriptsize$\,\pm\,0.17$} & $0.00${\scriptsize$\,\pm\,0.00$} & $0.0${\scriptsize$\,\pm\,0.1$} & 0.38{\scriptsize$\,\pm\,0.20$} & 2.51 & 0.12 & A10 \\
  & UnAct (ours) & \texttt{\scriptsize p90.0, g0.3, k1} & $76.56${\scriptsize$\,\pm\,0.27$} & $0.00${\scriptsize$\,\pm\,0.00$} & $0.0${\scriptsize$\,\pm\,0.0$} & 1.15{\scriptsize$\,\pm\,0.35$} & 2.78 & 2.18 & L4 \\
\bottomrule
\end{tabular}}
\end{table}

Table~\ref{tab:main-full} adds standard deviations, the Avg.\ Gap and the
fraction of modified parameters to Table~\ref{tab:main}. UnAct is not sparse: at
its selected configurations it modifies \numParamsCtenUnAct\%,
\numParamsCtwentyUnAct\% and \numParamsChundredUnAct\% of the parameters,
against \numParamsCtenSSD\%, \numParamsCtwentySSD\% and \numParamsChundredSSD\%
for SSD and \numParamsCtenLFSSD\%, \numParamsCtwentyLFSSD\% and
\numParamsChundredLFSSD\% for LFSSD.

\section{What distance to retraining shows, and what it misses}
\label{app:depth}

\begin{table}[t]
\centering
\caption{What $\Delta$ measures (ResNet-18, full forget set, mean over the five
forget classes, selected configurations). $A_f$: forget test accuracy.
Agr$_f$: top-1 agreement with the retrained model on forget-class test images,
i.e.\ whether the forgotten images go to the same classes as under retraining.
Probe: test AUROC of a class-balanced linear probe for the forget class, trained
on the model's own penultimate training features. The logit mask zeroes the
classifier row of the class and sets its bias to $-10^4$. The last two rows replay
UnAct's per-round channel masks on only the incoming ($\mathrm{In}_j$: last
block's \texttt{conv2} filters and BatchNorm affine) or only the outgoing
($\mathrm{Out}_j$: classifier columns) parameters. Evaluated on the A10.}
\label{tab:depth}
\small
\setlength{\tabcolsep}{2.4pt}
\begin{tabular}{@{}lcccccccccccc@{}}
\toprule
& \multicolumn{4}{c}{CIFAR-10} & \multicolumn{4}{c}{CIFAR-20} & \multicolumn{4}{c}{CIFAR-100} \\
\cmidrule(lr){2-5}\cmidrule(lr){6-9}\cmidrule(l){10-13}
Method & $A_f$ & $\Delta$ & Agr$_f$ & Probe & $A_f$ & $\Delta$ & Agr$_f$ & Probe & $A_f$ & $\Delta$ & Agr$_f$ & Probe \\
\midrule
Original model & 93.3 & 93.73 & 5.3 & 0.992 & 89.0 & 89.49 & 7.2 & 0.991 & 81.0 & 81.39 & 10.0 & 0.981 \\
Retrain & 0.0 & --- & 100.0 & 0.977 & 0.0 & --- & 100.0 & 0.972 & 0.0 & --- & 100.0 & 0.956 \\
\midrule
Logit mask (trivial) & 0.0 & 0.09 & 62.4 & 0.992 & 0.0 & 0.25 & 46.0 & 0.991 & 0.0 & 0.33 & 45.0 & 0.981 \\
\midrule
SSD & 0.0 & 0.29 & 65.5 & 0.992 & 0.0 & 0.26 & 48.8 & 0.990 & 0.0 & 0.51 & 39.6 & 0.979 \\
LFSSD & 0.0 & 0.11 & 62.5 & 0.991 & 0.0 & 0.24 & 47.4 & 0.990 & 0.0 & 0.38 & 43.2 & 0.979 \\
\midrule
\textbf{UnAct} & 0.0 & 0.08 & 31.4 & 0.991 & 0.0 & 0.79 & 28.4 & 0.989 & 0.0 & 1.14 & 27.8 & 0.985 \\
\quad $\mathrm{In}_j$ only & 51.7 & 51.82 & 25.5 & 0.991 & 21.2 & 21.89 & 28.8 & 0.989 & 20.6 & 21.49 & 29.6 & 0.985 \\
\quad $\mathrm{Out}_j$ only & 0.6 & 0.67 & 31.7 & 0.992 & 20.2 & 20.85 & 28.5 & 0.991 & 41.4 & 42.11 & 25.8 & 0.981 \\
\bottomrule
\end{tabular}
\end{table}

Every comparison in Section~\ref{sec:results} is in $\Delta$, which is computed
from output accuracies. Here we ask what it certifies, with a trivial baseline and
three measurements beyond the output (Table~\ref{tab:depth}).

\paragraph{A trivial logit mask matches $\Delta$.} A baseline that removes the
class logit and changes nothing else reaches $\Delta =$ \numDepthDeltaMaskCten,
\numDepthDeltaMaskCtwenty{} and \numDepthDeltaMaskChundred. This is comparable to
SSD and LFSSD, and lower than UnAct's on CIFAR-20 and CIFAR-100. It is also
unaffected by the forget-set size, since it needs the class index and no images
at all. $\Delta$ therefore shows that a class is no longer predicted and that
retain accuracy is preserved. For none of the methods does it show that the class
has been removed, and throughout this paper we use it only in that sense.

\paragraph{UnAct's edit is not simply a classifier edit.} Replaying UnAct's
per-round channel masks on the classifier columns alone nearly forgets the class
on CIFAR-10 (\numDepthDfUnActOutCten\% forget accuracy). On CIFAR-20 and CIFAR-100
it leaves forget accuracy at \numDepthDfUnActOutCtwenty\% and
\numDepthDfUnActOutChundred\%, and the incoming weights alone leave
\numDepthDfUnActInMin--\numDepthDfUnActInMax\%. Only the combination forgets on
every dataset.

\paragraph{No method removes the class from the features.} A linear probe
\citep{alain2017understanding} trained on each model's penultimate features
recovers the forget class with test AUROC of at least \numDepthProbeMin{} after
every method. The probe cannot rank the methods, because it recovers the class
almost as well from the retrained model (AUROC $\geq \numDepthProbeGoldMin$),
which never saw it: features learned from the other classes already separate it
linearly. Scaling a unit by $\gamma$ instead of zeroing it also leaves its signal
linearly recoverable, which is consistent with the fast relearning of every
method (Appendix~\ref{app:relearn}).

\paragraph{UnAct sends forgotten images to different classes than retraining
does.} SSD, LFSSD and the logit mask send \numDepthAgrFSSDCten\%,
\numDepthAgrFLFSSDCten\% and \numDepthAgrFMaskCten\% of the forget images on
CIFAR-10 to the class the retrained model predicts for them, against
\numDepthAgrFUnActCten\% for UnAct, and the pattern is the same on CIFAR-20 and
CIFAR-100. UnAct's attenuation, with BatchNorm statistics frozen, moves forget
images away from the retrained model's predictions. This counts against UnAct,
and $\Delta$ does not show it.

\paragraph{Details.} Table~\ref{tab:depth} was evaluated on the A10 GPU; recomputed there, the SSD,
LFSSD and UnAct rows reproduce the $\Delta$ of Table~\ref{tab:main} to within
0.01. The logit mask zeroes the
classifier row of the forget class and sets its bias to $-10^4$. The in/out
rows replay the channel masks recorded in each round of the full UnAct run on
only $\mathrm{In}_j$ (the last block's final convolution filters and BatchNorm
affine parameters) or only $\mathrm{Out}_j$ (the classifier columns), so all
three UnAct rows edit exactly the same channels. Agreement is the top-1
agreement with the retrained model on forget-class test images. The probe
standardises the 512 pooled features with training-set statistics and fits an
$\ell_2$-regularised logistic regression (forget class against the rest) by
full-batch L-BFGS on the model's own training features, with the positive class
re-weighted to balance the classes; we report its AUROC on the test features.
Because it is refit on each model's features, it measures whether the class is
still linearly decodable, not whether the original readout still works.
Standardisation undoes a uniform rescaling of a feature, so attenuation can lower
the probe's accuracy only through the non-uniform BatchNorm rescaling of
Appendix~\ref{app:proofs} and the ReLU and shortcut that follow the edited
branch. The logit mask and the $\mathrm{Out}_j$-only replay leave the
features unchanged, so their probe values equal the original model's.

\section{Forget-set size}
\label{app:data-efficiency}

\begin{figure}[t]
\begin{center}
\includegraphics[width=\linewidth]{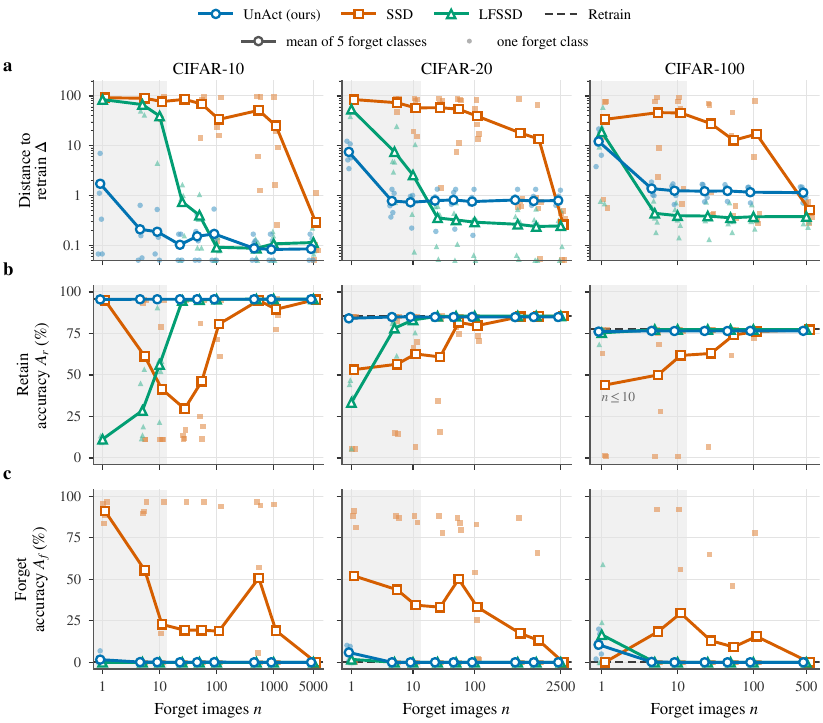}
\end{center}
\caption{As Figure~\ref{fig:forget-size}, for the three ResNet-18 datasets, with
$\Delta$ on a log scale (a) so that differences below one point are visible. Lines: mean
over the five forget classes; small faint markers: each class (slightly offset in
$n$ for legibility). SSD and LFSSD are bimodal at small $n$: on some classes
retain accuracy collapses and forget accuracy falls with it, while others are
barely changed. Values of $\Delta$ below 0.05 are drawn at 0.05.}
\label{fig:forget-size-full}
\end{figure}

\begin{table}[t]
\centering
\caption{Distance to the retrained model $\Delta$ against the number of forget
images $n$ each method is shown; the best method in each cell group is in bold.
Every method runs at its single selected configuration at every $n$. The class
being removed and the retain set are fixed, and SSD and LFSSD always receive
their importance pass over the full training set, so $n$ measures only how much
evidence a method needs to locate the class. The last $n$ of each dataset is the
whole class. $n^\star$ is the smallest $n$ at which a method comes within
$2\times$ of its own best $\Delta$.}
\label{tab:data-efficiency}
\small
\setlength{\tabcolsep}{3.5pt}
\begin{tabular}{@{}rccccccccc@{}}
\toprule
& \multicolumn{3}{c}{CIFAR-10} & \multicolumn{3}{c}{CIFAR-20} & \multicolumn{3}{c}{CIFAR-100} \\
\cmidrule(lr){2-4}\cmidrule(lr){5-7}\cmidrule(lr){8-10}
$n$ & SSD & LFSSD & UnAct & SSD & LFSSD & UnAct & SSD & LFSSD & UnAct \\
\midrule
1 & 91.85 & 84.34 & \textbf{1.72} & 84.53 & 54.05 & \textbf{7.50} & 33.79 & 18.97 & \textbf{12.19} \\
5 & 89.85 & 67.07 & \textbf{0.21} & 72.95 & 7.55 & \textbf{0.77} & 46.14 & \textbf{0.44} & 1.37 \\
10 & 77.02 & 39.40 & \textbf{0.19} & 57.46 & 2.56 & \textbf{0.73} & 45.57 & \textbf{0.39} & 1.24 \\
25 & 85.35 & 0.74 & \textbf{0.10} & 58.06 & \textbf{0.35} & 0.79 & 27.77 & \textbf{0.39} & 1.22 \\
50 & 68.73 & 0.40 & \textbf{0.15} & 54.16 & \textbf{0.32} & 0.81 & 12.85 & \textbf{0.36} & 1.23 \\
100 & 33.76 & \textbf{0.09} & 0.17 & 39.23 & \textbf{0.29} & 0.75 & 17.10 & \textbf{0.37} & 1.16 \\
500 & 51.39 & \textbf{0.09} & \textbf{0.09} & 18.02 & \textbf{0.26} & 0.81 & 0.51 & \textbf{0.38} & 1.14 \\
1000 & 25.12 & 0.11 & \textbf{0.08} & 13.61 & \textbf{0.24} & 0.77 & --- & --- & --- \\
2500 & --- & --- & --- & 0.26 & \textbf{0.24} & 0.79 & --- & --- & --- \\
5000 & 0.29 & 0.11 & \textbf{0.08} & --- & --- & --- & --- & --- & --- \\
\midrule
$n^\star$ & 5000 & 100 & 25 & 2500 & 25 & 5 & 500 & 5 & 5 \\
\bottomrule
\end{tabular}
\end{table}

Figure~\ref{fig:forget-size-full} shows every run of the forget-set-size sweep and
Table~\ref{tab:data-efficiency} every mean. The smallest $n$ at which a method
comes within a factor of two of its own best $\Delta$ is
$n^\star = \numNstarUnActCten$, $\numNstarUnActCtwenty$ and
$\numNstarUnActChundred$ for UnAct, $\numNstarLFSSDCten$, $\numNstarLFSSDCtwenty$
and $\numNstarLFSSDChundred$ for LFSSD, and the whole class
($\numNstarSSDCten$, $\numNstarSSDCtwenty$ and $\numNstarSSDChundred$) for SSD.
Because $n^\star$ is relative to each method's own best, it should be read
together with the absolute values in Table~\ref{tab:data-efficiency}.

\begin{table}[t]
\centering
\caption{SSD with $\alpha$ re-selected at every forget-set size (ResNet-18,
$\Delta$, mean over the five forget classes). UnAct and SSD run at their
full-class configurations, as in Figure~\ref{fig:forget-size}. SSD-$n$ picks, at
each $n$ separately, the $\alpha$ in the range reported by the SSD paper
($\alpha\in[5,50]$; value in parentheses) with the lowest mean $\Delta$, which
requires a retrained model at every $n$; $\lambda$ and the importance batch stay at
their selected values. UnAct receives no per-$n$ tuning. From the existing
forget-set-size runs; ``all'' is the whole class (500 images on CIFAR-100).}
\label{tab:ssd-oracle}
\small
\setlength{\tabcolsep}{3pt}
\begin{tabular}{@{}rccccccccc@{}}
\toprule
& \multicolumn{3}{c}{CIFAR-10} & \multicolumn{3}{c}{CIFAR-20} & \multicolumn{3}{c}{CIFAR-100} \\
\cmidrule(lr){2-4}\cmidrule(lr){5-7}\cmidrule(l){8-10}
$n$ & UnAct & SSD & SSD-$n$ & UnAct & SSD & SSD-$n$ & UnAct & SSD & SSD-$n$ \\
\midrule
1 & 1.72 & 91.85 & 89.26\,{\scriptsize(6)} & 7.50 & 84.53 & 83.68\,{\scriptsize(6)} & 12.19 & 33.79 & 33.79\,{\scriptsize(50)} \\
5 & 0.21 & 89.85 & 87.47\,{\scriptsize(6)} & 0.77 & 72.95 & 64.62\,{\scriptsize(50)} & 1.37 & 46.14 & 46.14\,{\scriptsize(50)} \\
10 & 0.19 & 77.02 & 77.02\,{\scriptsize(8)} & 0.73 & 57.46 & 55.91\,{\scriptsize(20)} & 1.24 & 45.57 & 25.17\,{\scriptsize(15)} \\
25 & 0.10 & 85.35 & 77.46\,{\scriptsize(20)} & 0.79 & 58.06 & 57.13\,{\scriptsize(6)} & 1.22 & 27.77 & 16.04\,{\scriptsize(30)} \\
50 & 0.15 & 68.73 & 65.45\,{\scriptsize(10)} & 0.81 & 54.16 & 38.42\,{\scriptsize(12)} & 1.23 & 12.85 & 7.39\,{\scriptsize(30)} \\
100 & 0.17 & 33.76 & 31.20\,{\scriptsize(10)} & 0.75 & 39.23 & 22.48\,{\scriptsize(10)} & 1.16 & 17.10 & 5.31\,{\scriptsize(30)} \\
500 & 0.09 & 51.39 & 8.91\,{\scriptsize(6)} & 0.81 & 18.02 & 1.83\,{\scriptsize(10)} &  &  &  \\
1000 & 0.08 & 25.12 & 23.03\,{\scriptsize(6)} & 0.77 & 13.61 & 0.72\,{\scriptsize(12)} &  &  &  \\
all & 0.08 & 0.29 & 0.29\,{\scriptsize(8)} & 0.79 & 0.26 & 0.26\,{\scriptsize(15)} & 1.14 & 0.51 & 0.51\,{\scriptsize(50)} \\
\bottomrule
\end{tabular}
\end{table}

Table~\ref{tab:ssd-oracle} separates two explanations of SSD's failure at small
$n$: that SSD cannot work from a few images, and that its full-class $\alpha$
does not transfer to them. With $\alpha$ re-selected at each $n$, SSD improves
substantially at several sizes, most clearly at $n=500$ (8.91 against 51.39 on
CIFAR-10 and 1.83 against 18.02 on CIFAR-20), but not uniformly: on CIFAR-10 at
$n=1000$ it stays at 23.03 against 25.12. At $n \leq 10$, though, its
$\Delta$ stays at \numOracleSSDSmallMin{} or above on every dataset, while
UnAct's single configuration stays near its full-class value. Within the
published range, SSD's small-$n$ failure is therefore not only a calibration
artefact of our protocol. The oracle is available only for SSD, because the
forget-set-size runs replayed its $\alpha$ grid but not LFSSD's.

\section{Several classes}
\label{app:multiclass}

\begin{figure}[t]
\begin{center}
\includegraphics[width=\linewidth]{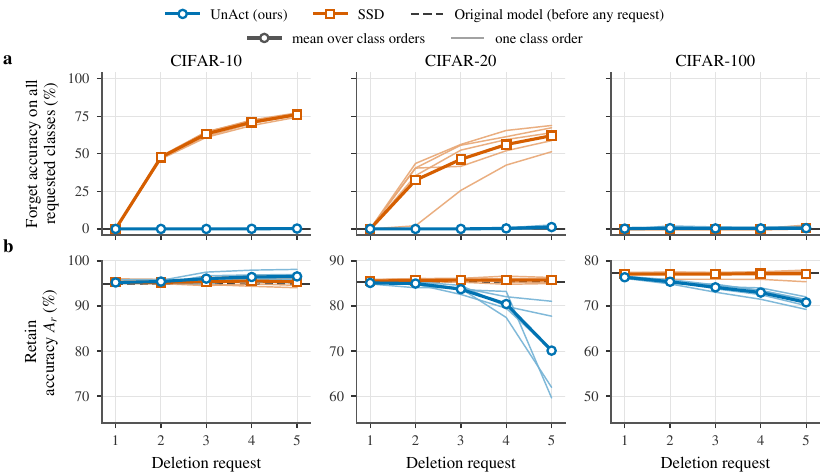}
\end{center}
\caption{\textbf{Sequential deletion requests.} Five one-class requests applied
one after another to the same model, each method at its selected single-class
configuration with no re-tuning. (a) Forget accuracy on all classes requested so
far. (b) Retain accuracy. Thick lines: mean over five class orders; thin lines:
individual orders. The dashed line is the original model's test accuracy before
any request, not a retrained model: most prefixes have no retrained reference.
Per-request values are in Table~\ref{tab:sequential}.}
\label{fig:sequential}
\end{figure}

Figure~\ref{fig:sequential} applies five one-class requests in sequence, each to
the already-edited model, for five class orders per dataset drawn at random before
running. Each method runs at its selected single-class configuration, and SSD
recomputes its importance on the current model at every request. LFSSD was not run
in this experiment.

\paragraph{On CIFAR-10 and CIFAR-20, UnAct keeps every requested class
forgotten; SSD does not.} In every class order, SSD's forget accuracy on the
requested classes is above 50\% by the fifth request: \numSeqDfSSDCtenRtwo\% and
\numSeqDfSSDCtwentyRtwo\% after the second request and \numSeqDfSSDCtenRfive\%
and \numSeqDfSSDCtwentyRfive\% after the fifth (between
\numSeqDfMinOrderSSDCtwentyRfive\% and \numSeqDfMaxOrderSSDCtwentyRfive\% across
CIFAR-20 orders). Across all orders and requests, UnAct's forget accuracy on the
requested classes never exceeds \numSeqDfMaxUnActCtenCtwenty\%. On CIFAR-10,
UnAct's retain accuracy does not erode. It rises from \numSeqDrUnActCtenRone\% to
\numSeqDrUnActCtenRfive\% as classes are removed, as a retrained model's does. On
CIFAR-20 it is stable for about three requests (\numSeqDrUnActCtwentyRone\% to
\numSeqDrUnActCtwentyRthree\%) and then erodes, to \numSeqDrUnActCtwentyRfive\%
after the fifth (worst order \numSeqDrWorstUnActCtwentyRfive\%), while SSD's
stays at \numSeqDrSSDCtwentyRfive\%.

\paragraph{On CIFAR-100, SSD is the better choice.} SSD keeps forgetting (forget
accuracy at most \numSeqDfSSDChundredRfive\% on average) and keeps its retain
accuracy at \numSeqDrSSDChundredRfive\%. UnAct also forgets but loses about
\numSeqDrPerRequestUnActChundred{} points of retain accuracy per request
(\numSeqDrUnActChundredRone\% to \numSeqDrUnActChundredRfive\%). Two
measurements point to the cause. First, a single UnAct request on CIFAR-100
already costs \numDrGapUnActChundred{} points of retain accuracy
(Table~\ref{tab:main}), and these costs add up over requests, while SSD edits
roughly ten times fewer parameters (\numParamsChundredSSD\% against
\numParamsChundredUnAct\%). Second, the more classes there are, the more of them
share each of the 512 channels of the last residual block, so the channels most
active for one class are less specific to it. At a fixed percentile, the ratio of
forget-set to retain-set activation carried by the selected channels falls from
\numSelRatioCten{} on CIFAR-10 to \numSelRatioChundred{} on CIFAR-100
(Appendix~\ref{app:selectivity}). Removing several classes in a single request,
at the single-class settings, fails for both methods
(Table~\ref{tab:multiclass}); UnAct is designed for one class per request.

\begin{table}[t]
\centering
\caption{Sequential deletion requests over five class orders per dataset, drawn
at random before running
(one class per request, each applied to the already-edited model). Mean
$\pm$ sd over the class orders of the forget accuracy on \emph{all} classes
requested so far ($A_f$) and the retain accuracy ($A_r$), both in \%. Both
methods run at their selected single-class configuration, with no re-tuning;
SSD recomputes its importances on the current model at every request. Most
prefixes have no retrained model, so the reference is the original model's
test accuracy over all classes (``orig.''); its accuracy on the remaining classes
shifts as classes are removed. The first order is the one of
Table~\ref{tab:multiclass} (evaluated on an A10); the others were evaluated on an
L4. Seed 42.}
\label{tab:sequential}
\scriptsize
\setlength{\tabcolsep}{3.2pt}
\begin{tabular}{@{}llcccc@{}}
\toprule
& & \multicolumn{2}{c}{SSD} & \multicolumn{2}{c}{UnAct (ours)} \\
\cmidrule(lr){3-4}\cmidrule(l){5-6}
Dataset & Request & $A_f$ & $A_r$ & $A_f$ & $A_r$ \\
\midrule
\multirow{5}{*}{\shortstack[l]{CIFAR-10\\\scriptsize 5 orders\\\scriptsize orig.\ 94.97}} & 1 & $0.0${\scriptsize$\,\pm\,0.0$} & $95.3${\scriptsize$\,\pm\,0.4$} & $0.0${\scriptsize$\,\pm\,0.0$} & $95.2${\scriptsize$\,\pm\,0.3$} \\
 & 2 & $47.7${\scriptsize$\,\pm\,0.7$} & $95.2${\scriptsize$\,\pm\,0.4$} & $0.0${\scriptsize$\,\pm\,0.0$} & $95.4${\scriptsize$\,\pm\,0.4$} \\
 & 3 & $63.3${\scriptsize$\,\pm\,1.8$} & $95.4${\scriptsize$\,\pm\,0.7$} & $0.0${\scriptsize$\,\pm\,0.0$} & $96.1${\scriptsize$\,\pm\,1.0$} \\
 & 4 & $71.0${\scriptsize$\,\pm\,1.5$} & $95.5${\scriptsize$\,\pm\,0.9$} & $0.1${\scriptsize$\,\pm\,0.1$} & $96.4${\scriptsize$\,\pm\,1.0$} \\
 & 5 & $76.1${\scriptsize$\,\pm\,1.1$} & $95.3${\scriptsize$\,\pm\,1.1$} & $0.3${\scriptsize$\,\pm\,0.3$} & $96.6${\scriptsize$\,\pm\,1.1$} \\
\multicolumn{6}{@{}l}{\tiny orders: (0,2,3,5,8); (5,6,1,2,0); (8,7,1,5,6); (6,0,3,7,8); (0,4,9,6,7)} \\
\midrule
\multirow{5}{*}{\shortstack[l]{CIFAR-20\\\scriptsize 5 orders\\\scriptsize orig.\ 85.34}} & 1 & $0.0${\scriptsize$\,\pm\,0.0$} & $85.5${\scriptsize$\,\pm\,0.3$} & $0.0${\scriptsize$\,\pm\,0.1$} & $85.1${\scriptsize$\,\pm\,0.2$} \\
 & 2 & $32.5${\scriptsize$\,\pm\,17.2$} & $85.6${\scriptsize$\,\pm\,0.4$} & $0.0${\scriptsize$\,\pm\,0.0$} & $85.0${\scriptsize$\,\pm\,0.6$} \\
 & 3 & $46.4${\scriptsize$\,\pm\,13.0$} & $85.7${\scriptsize$\,\pm\,0.3$} & $0.0${\scriptsize$\,\pm\,0.0$} & $83.7${\scriptsize$\,\pm\,0.7$} \\
 & 4 & $56.2${\scriptsize$\,\pm\,9.1$} & $85.7${\scriptsize$\,\pm\,0.7$} & $0.4${\scriptsize$\,\pm\,0.4$} & $80.4${\scriptsize$\,\pm\,2.3$} \\
 & 5 & $62.1${\scriptsize$\,\pm\,7.1$} & $85.7${\scriptsize$\,\pm\,0.6$} & $1.2${\scriptsize$\,\pm\,1.0$} & $70.1${\scriptsize$\,\pm\,9.4$} \\
\multicolumn{6}{@{}l}{\tiny orders: (3,4,10,14,19); (5,13,2,11,19); (8,1,19,0,10); (6,12,3,7,1); (10,14,15,2,5)} \\
\midrule
\multirow{5}{*}{\shortstack[l]{CIFAR-100\\\scriptsize 5 orders\\\scriptsize orig.\ 77.35}} & 1 & $0.0${\scriptsize$\,\pm\,0.0$} & $77.1${\scriptsize$\,\pm\,0.6$} & $0.2${\scriptsize$\,\pm\,0.4$} & $76.4${\scriptsize$\,\pm\,0.2$} \\
 & 2 & $0.0${\scriptsize$\,\pm\,0.0$} & $77.1${\scriptsize$\,\pm\,0.7$} & $0.5${\scriptsize$\,\pm\,0.9$} & $75.4${\scriptsize$\,\pm\,0.4$} \\
 & 3 & $0.1${\scriptsize$\,\pm\,0.1$} & $77.1${\scriptsize$\,\pm\,0.6$} & $0.3${\scriptsize$\,\pm\,0.6$} & $74.1${\scriptsize$\,\pm\,0.7$} \\
 & 4 & $0.1${\scriptsize$\,\pm\,0.1$} & $77.2${\scriptsize$\,\pm\,0.7$} & $0.3${\scriptsize$\,\pm\,0.5$} & $73.0${\scriptsize$\,\pm\,1.0$} \\
 & 5 & $0.5${\scriptsize$\,\pm\,1.2$} & $77.1${\scriptsize$\,\pm\,1.0$} & $0.5${\scriptsize$\,\pm\,0.6$} & $70.8${\scriptsize$\,\pm\,1.1$} \\
\multicolumn{6}{@{}l}{\tiny orders: (3,20,51,69,85); (45,15,90,32,35); (48,97,1,81,90); (86,42,89,92,3); (30,13,49,91,37)} \\
\bottomrule
\end{tabular}
\end{table}

\begin{table}[t]
\centering
\caption{Removing several classes, with each method at its selected single-class
configuration (no re-tuning for the task). \emph{Sequential}: one class per
request, applied to the already-edited model; the column is the number of
requests so far. \emph{Simultaneous}: all $q$ classes in one request. $\Delta$ is
measured against a model retrained without all $q$ classes; four-class retrained
models were not trained, so $q=4$ is omitted. The last column of each method is
$A_r/A_f$ after five classes. Sequential: the first of the five class orders of
Table~\ref{tab:sequential}, the only one with retrained references. Seed 42.}
\label{tab:multiclass}
\scriptsize
\setlength{\tabcolsep}{2.6pt}
\begin{tabular}{@{}llccccrccccr@{}}
\toprule
& & \multicolumn{5}{c}{SSD} & \multicolumn{5}{c}{UnAct (ours)} \\
\cmidrule(lr){3-7}\cmidrule(l){8-12}
& Dataset & $q{=}1$ & 2 & 3 & 5 & $A_r/A_f$ & $q{=}1$ & 2 & 3 & 5 & $A_r/A_f$ \\
\midrule
\multirow{3}{*}{\rotatebox{90}{\scriptsize sequential}} & CIFAR-10 & 0.00 & 46.67 & 62.21 & 75.60 & 96.8/74.4 & 0.00 & 0.16 & 0.13 & 0.60 & 98.1/0.4 \\
 & CIFAR-20 & 0.43 & 43.73 & 56.37 & 69.07 & 85.0/68.9 & 0.63 & 0.39 & 2.84 & 15.21 & 70.1/0.2 \\
 & CIFAR-100 & 0.72 & 0.46 & 0.76 & 2.79 & 77.3/2.6 & 1.48 & 4.59 & 4.17 & 7.63 & 71.4/1.6 \\
\midrule
\multirow{3}{*}{\rotatebox{90}{\scriptsize simultaneous}} & CIFAR-10 & 0.00 & 94.80 & 93.67 & 94.54 & 96.7/93.3 & 0.00 & 55.29 & 72.41 & 84.76 & 98.4/84.3 \\
 & CIFAR-20 & 0.43 & 86.83 & 88.74 & 89.97 & 84.1/89.0 & 0.63 & 13.64 & 31.10 & 60.51 & 85.3/60.4 \\
 & CIFAR-100 & 0.72 & 46.24 & 77.98 & 81.29 & 77.2/81.0 & 1.48 & 10.97 & 21.44 & 43.66 & 77.0/43.2 \\
\bottomrule
\end{tabular}
\end{table}

Table~\ref{tab:sequential} gives the per-request values behind
Figure~\ref{fig:sequential}. For the first class order, retrained models exist
for most prefixes, so Table~\ref{tab:multiclass} also reports $\Delta$; after the
fifth request on CIFAR-10 the retrained model's retain accuracy is
\numSequentialGoldDrCtenKfive\%, which is why UnAct's rising retain accuracy there
is expected. Table~\ref{tab:multiclass} also removes $q$ classes in a single
request at the single-class settings. Both methods fail at this task for $q>1$,
UnAct less badly than SSD ($\Delta$ of \numSimultaneousUnActCtenKfive{},
\numSimultaneousUnActCtwentyKfive{} and \numSimultaneousUnActChundredKfive{}
against \numSimultaneousSSDCtenKfive{}, \numSimultaneousSSDCtwentyKfive{} and
\numSimultaneousSSDChundredKfive{} at $q=5$). Neither was tuned for it.

\section{Class selectivity}
\label{app:selectivity}

For each of the 512 output channels of the last residual block we compute $\rho =
(a_f - a_r)/(a_f + a_r)$, where $a_f$ and $a_r$ are the channel's mean post-ReLU
activations over the forget-class and retain training images; no weight is
modified. Selectivity lies in $[-1,1]$: $\rho=1$ means the channel responds to the
forget class alone, and $\rho=0$ means it responds equally to both, so attenuating
it removes as much retain signal as forget signal. We estimate $a_r$ on a sample
of the retain training set and have not quantified the resulting sampling error.

\begin{figure}[t]
\begin{center}
\includegraphics[width=\linewidth]{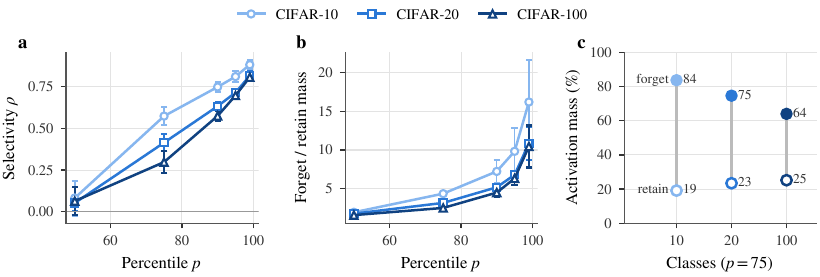}
\end{center}
\caption{Class selectivity of the selected channels, from forward passes only.
Error bars are one standard deviation over the five forget classes. (a)
Selectivity rises with the percentile $p$. (b) The forget-to-retain activation
mass ratio of the selected channels. (c) At fixed $p=75$, as more classes share
the same 512 channels, the selected channels carry less forget mass and more
retain mass.}
\label{fig:selectivity}
\end{figure}

Selection by activation alone lands on class-specific channels: the selected
channels become sharply more selective as the percentile $p$ rises, although the
rule never consults the retain set (Figure~\ref{fig:selectivity}a). At a fixed
$p = 75$, selectivity falls as the same 512 channels are shared by more classes
(Table~\ref{tab:selectivity}): the selected channels capture less of the forget
set's activation and more of the retain set's, and the ratio of the two falls
from \numSelRatioCten{} to \numSelRatioCtwenty{} to \numSelRatioChundred{}. This
is the measurement Appendix~\ref{app:multiclass} uses. It does not predict the
selected percentiles, which are 99, 99 and 90: $p$ and $k$ trade off, so the
amount of the layer that is edited is better described by the coverage
$\Gamma_k$ (Appendix~\ref{app:coverage}).

\begin{table}[t]
\centering
\caption{Class selectivity of the selected channels at a fixed $p=75$, measured
on the $512$-channel output of the last residual block. These are forward-pass
statistics and no weight is modified. Per channel, $\rho=(a_f-a_r)/(a_f+a_r)$,
where $a_f$ and $a_r$ are mean activations on the forget and retain sets. Forget
and retain mass are the fractions of each set's total activation that the
selected channels carry. Mean $\pm$ s.d.\ over the five forget classes.}
\label{tab:selectivity}
\small
\begin{tabular}{@{}lccccc@{}}
\toprule
Dataset & Classes & Selectivity $\rho$ & Forget mass (\%) & Retain mass (\%) & Ratio \\
\midrule
CIFAR-10 & 10 & $0.574${\scriptsize$\,\pm\,0.053$} & $83.7${\scriptsize$\,\pm\,3.5$} & $19.1${\scriptsize$\,\pm\,0.5$} & $4.38${\scriptsize$\,\pm\,0.24$} \\
CIFAR-20 & 20 & $0.414${\scriptsize$\,\pm\,0.052$} & $74.6${\scriptsize$\,\pm\,0.5$} & $23.4${\scriptsize$\,\pm\,0.4$} & $3.19${\scriptsize$\,\pm\,0.03$} \\
CIFAR-100 & 100 & $0.298${\scriptsize$\,\pm\,0.066$} & $64.0${\scriptsize$\,\pm\,4.6$} & $25.2${\scriptsize$\,\pm\,0.3$} & $2.54${\scriptsize$\,\pm\,0.16$} \\
\bottomrule
\end{tabular}
\end{table}

\section{Cost}
\label{app:cost}

\begin{figure}[t]
\begin{center}
\includegraphics[width=\linewidth]{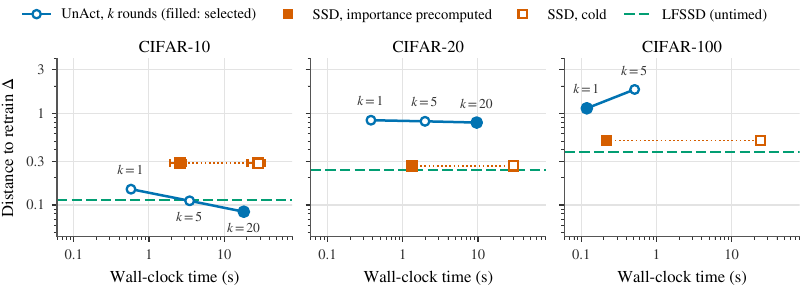}
\end{center}
\caption{\textbf{Cost of one request} (full forget class, class 3, one L4 GPU).
Distance to retraining against wall-clock time. UnAct runs the best configuration
of its grid for $k=1$ and $k=5$ rounds and its selected configuration (filled).
SSD runs its selected configuration with its full-training-set importance
precomputed (filled) or computed at request time (cold, open); whiskers span the
timing sessions. LFSSD's $\Delta$ is drawn as a line; it was not timed and has
SSD's two-pass structure. Paired per-session timings are in
Table~\ref{tab:cost}.}
\label{fig:cost}
\end{figure}

\begin{table}[t]
\centering
\caption{Wall-clock cost on one L4, serialised, full forget class (class 3),
warm-up repeat discarded, mean of the remaining repeats. For each $k$, UnAct runs
the best configuration of its grid with that number of rounds
($^\dagger$: the selected configuration); its $\Delta$ is the five-class grid
value. SSD runs its selected configuration in the same session, so each ratio is
paired. ``Amortised'' credits SSD's importance pass over the full training set as
precomputed and reusable, which its paper allows; that pass is the listed share
of SSD's cold cost. Ratios above 1 mean UnAct is faster; bold marks the
settings where it is faster even than amortised SSD.}
\label{tab:cost}
\small
\setlength{\tabcolsep}{3.2pt}
\begin{tabular}{@{}llccccccrr@{}}
\toprule
& & & & UnAct & \multicolumn{2}{c}{SSD (s)} & full-$D$ & \multicolumn{2}{c}{UnAct speed-up} \\
\cmidrule(lr){6-7}\cmidrule(l){9-10}
Dataset & $k$ & config & $\Delta$ & (s) & amortised & cold & share (\%) & vs.\ amort. & vs.\ cold \\
\midrule
\multirow{3}{*}{CIFAR-10} & 1 & \texttt{\scriptsize p75, $\gamma$0.01} & 0.15 & 0.58 & 1.91 & 20.4 & 91 & \textbf{3.31}$\times$ & 35$\times$ \\
 & 5 & \texttt{\scriptsize p90, $\gamma$0.03} & 0.11 & 3.48 & 2.72 & 30.5 & 91 & 0.78$\times$ & 9$\times$ \\
 & 20$^\dagger$ & \texttt{\scriptsize p99, $\gamma$0.01} & 0.08 & 18.27 & 3.08 & 34.5 & 91 & 0.17$\times$ & 2$\times$ \\
\midrule
\multirow{3}{*}{CIFAR-20} & 1 & \texttt{\scriptsize p75, $\gamma$0.1} & 0.84 & 0.38 & 1.14 & 26.3 & 96 & \textbf{3.01}$\times$ & 69$\times$ \\
 & 5 & \texttt{\scriptsize p95, $\gamma$0.1} & 0.82 & 1.98 & 1.31 & 29.2 & 95 & 0.66$\times$ & 15$\times$ \\
 & 20$^\dagger$ & \texttt{\scriptsize p99, $\gamma$0.1} & 0.80 & 9.64 & 1.53 & 33.9 & 95 & 0.16$\times$ & 4$\times$ \\
\midrule
\multirow{2}{*}{CIFAR-100} & 1$^\dagger$ & \texttt{\scriptsize p90, $\gamma$0.3} & 1.14 & 0.12 & 0.22 & 25.5 & 99 & \textbf{1.83}$\times$ & 215$\times$ \\
 & 5 & \texttt{\scriptsize p99, $\gamma$0.03} & 1.83 & 0.51 & 0.22 & 22.8 & 99 & 0.42$\times$ & 45$\times$ \\
\bottomrule
\end{tabular}
\end{table}

Figure~\ref{fig:cost} plots the trade-off and Table~\ref{tab:cost} lists the
timings. Each UnAct
setting was timed in the same session as SSD, so each ratio is paired. SSD's time
for identical work varies between sessions, and the $k=1$ session recorded the
fastest SSD, so the one-round ratios quoted in Section~\ref{sec:cost} are
conservative. At $k = 5$ on every dataset, and at the selected $k=20$ on
CIFAR-10 and CIFAR-20, UnAct is slower than SSD with precomputed importance and
faster than SSD from cold.

\section{Vision transformer details}
\label{app:vit}

\begin{table}[t]
\centering
\caption{Removing one class from ViT-B/16 on CIFAR-10, mean over the five
forget classes. UnAct uses the class-token score (Section~\ref{sec:scoring};
$p=85$, $\gamma=0.1$, $k=5$, chosen on class 3); SSD uses its best grid setting
($\alpha=5$, $\lambda=0.1$, importance batch 128). Worst $A_r$: lowest retain
accuracy over the five classes; closer: classes on which the method has the
smaller $\Delta$. Per-class results are in Table~\ref{tab:vit-variants}.}
\label{tab:vit}
\small
\setlength{\tabcolsep}{4pt}
\begin{tabular}{@{}lcccccc@{}}
\toprule
Method & $A_r$ & $A_f$ & $\Delta\!\downarrow$ & worst $A_r$ & closer & Params (\%) \\
\midrule
Original model & 98.33 & 97.92 & 98.11 & 98.2 & --- & --- \\
Retrain & 98.46 & 0.00 & --- & 98.1 & --- & --- \\
\midrule
SSD & 78.89 & 14.16 & 33.74 & 9.6 & 2/5 & 6.44 \\
\textbf{UnAct} (ours) & 91.93 & 4.96 & \textbf{11.49} & 85.3 & 3/5 & 9.75 \\
\bottomrule
\end{tabular}
\end{table}

\begin{figure}[t]
\begin{center}
\includegraphics[width=\linewidth]{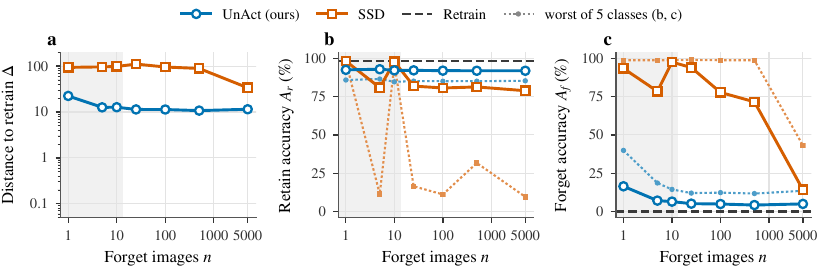}
\end{center}
\caption{Forget-set size on ViT-B/16 (CIFAR-10), UnAct and SSD at their selected
ViT configurations at every $n$, fp32 on the full test split. (a) $\Delta$, mean
over the five forget classes. (b) Retain accuracy: solid, mean; dotted, the
lowest of the five classes; dashed, the retrained models. (c) Forget accuracy:
solid, mean; dotted, the highest of the five classes; dashed, the retrained
models. Below the whole class, SSD leaves its worst class almost entirely
unforgotten, while UnAct's mean forget accuracy is below 10\% from $n=5$.}
\label{fig:vit-forget-size}
\end{figure}

\begin{table}[t]
\centering
\caption{ViT-B/16 score variants. Each UnAct variant's configuration
was chosen on class 3 (bf16, 2{,}000 retain images) and then run on all five
classes at fp32 on the full test split; the rows here are that second stage.
Columns 7--11 are $\Delta$ per forget class. Scoring by the positive part of the
activation at the class token fails (row 3), which is the reason for the logit score of
Section~\ref{sec:scoring}. SSD per class ($A_r/A_f$): 0: 98.1/43.2, 2: 98.2/1.6, 3: 96.1/23.1, 5: 92.5/2.9, 8: 9.6/0.0.}
\label{tab:vit-variants}
\scriptsize
\setlength{\tabcolsep}{2.4pt}
\resizebox{\textwidth}{!}{%
\begin{tabular}{@{}lllccccccccc@{}}
\toprule
& Units and score & Config & $A_r$ & $A_f$ & $\Delta$ & c0 & c2 & c3 & c5 & c8 & Params (\%) \\
\midrule
UnAct & MLP, blocks 9--12, class-token logit score (reported) & p85, $\gamma$0.1, $k$5 & 91.93 & 4.96 & 11.49 & 4.6 & 13.1 & 3.0 & 16.1 & 20.7 & 9.753 \\
UnAct & MLP, blocks 11--12, class-token logit score & p75, $\gamma$0.1, $k$5 & 95.45 & 41.22 & 44.24 & 50.7 & 33.1 & 22.5 & 63.7 & 51.1 & 5.558 \\
UnAct & MLP, blocks 9--12, class-token positive activation & p90, $\gamma$0.1, $k$10 & 94.48 & 91.42 & 95.41 & 102.9 & 99.6 & 66.7 & 106.1 & 101.8 & 7.210 \\
\midrule
SSD & importance-weighted dampening & $\alpha$5, $\lambda$0.1, b128 & 78.89 & 14.16 & 33.74 & 43.4 & 1.8 & 25.8 & 9.1 & 88.5 & 6.437 \\
\bottomrule
\end{tabular}}
\end{table}

Table~\ref{tab:vit-variants} compares UnAct's ViT score and scope variants, each
with its configuration chosen on class 3 and then run on all five classes. The
reported variant scores the MLP neurons of the last four blocks with
the class-token score of Section~\ref{sec:scoring}. Restricting it to the last two blocks raises $\Delta$ to
\numVitVariantFfnDlaLasttwo{}, and scoring the same neurons by the positive part of their
activation at the class token, the closest analogue of the ResNet score, fails ($\Delta =
\numVitVariantFfnPosLastfour$).

The class-token score of Section~\ref{sec:scoring} is computed in the forward
pass, but it reads the outgoing weights and the classifier row, much as a
gradient-times-activation score would. On ViT-B/16, UnAct is therefore
gradient-free in the sense of needing no backward pass, not in the sense of
using activations alone.

\section{Relearning}
\label{app:relearn}

\begin{table}[t]
\centering
\caption{Forget accuracy (\%) after 10 relearning steps, each on $m$ forget
images mixed $4{:}1$ with retain images (SGD, lr 0.01), mean over the five forget
classes. The retrained model is the reference: it never saw the class, so any
forget accuracy it gains is learned from the $m$ images alone. ``Ret.\ $A_r$'' is
the retrained model's own retain accuracy at step 10; at $m=1$ the probe degrades
it substantially, so small-$m$ curves mix relearning with damage
(Figure~\ref{fig:relearn-curves}).}
\label{tab:relearn}
\scriptsize
\setlength{\tabcolsep}{3pt}
\begin{tabular}{@{}lcccccccccccc@{}}
\toprule
& \multicolumn{4}{c}{$m=1$} & \multicolumn{4}{c}{$m=5$} & \multicolumn{4}{c}{$m=25$} \\
\cmidrule(lr){2-5}\cmidrule(lr){6-9}\cmidrule(l){10-13}
Dataset & Retrain & SSD & UnAct & Ret.\ $A_r$ & Retrain & SSD & UnAct & Ret.\ $A_r$ & Retrain & SSD & UnAct & Ret.\ $A_r$ \\
\midrule
CIFAR-10 & 0.0 & 47.2 & 25.9 & 68.4 & 0.0 & 68.1 & 36.8 & 87.5 & 0.0 & 87.4 & 47.5 & 92.7 \\
CIFAR-20 & 0.0 & 28.3 & 12.0 & 67.4 & 0.0 & 41.3 & 33.2 & 79.6 & 0.0 & 52.2 & 59.7 & 83.5 \\
CIFAR-100 & 1.0 & 53.2 & 54.0 & 63.8 & 1.2 & 40.2 & 83.6 & 71.8 & 6.6 & 52.2 & 95.2 & 75.2 \\
\bottomrule
\end{tabular}
\end{table}

\begin{figure}[t]
\begin{center}
\includegraphics[width=\linewidth]{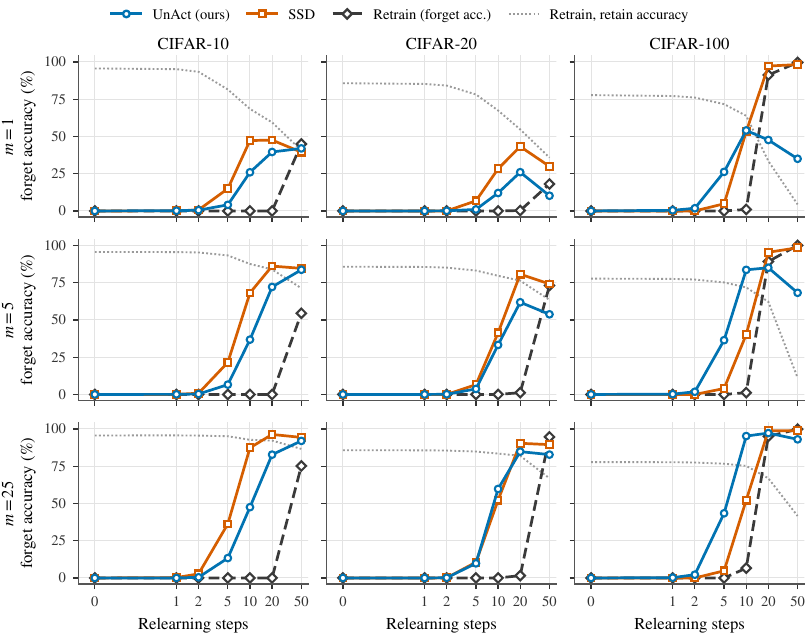}
\end{center}
\caption{Forget accuracy against gradient steps of relearning on $m$ forget
images per step, mixed $4{:}1$ with retain images (SGD, learning rate 0.01),
ResNet-18. The dotted line is the retrained model's own retain accuracy: at $m=1$
the probe degrades every model, so the late steps at $m=1$ mix relearning with
damage. Both unlearned models regain much of the class within 10 to 20 steps,
while the retrained model is still at or near zero forget accuracy at step 10.}
\label{fig:relearn-curves}
\end{figure}

A relearning probe fine-tunes each unlearned model on $m$ forget images per step,
mixed $4{:}1$ with retain images so that recovery cannot come from collapsing
onto the forget class, and tracks forget accuracy (Table~\ref{tab:relearn},
Figure~\ref{fig:relearn-curves}). Neither method deletes the class in a strong
sense: both recover it far faster than a retrained model learns it from the same
images. The order between the two is not consistent. At $m=5$, UnAct recovers
more slowly than SSD on CIFAR-10 and CIFAR-20 and faster on CIFAR-100, and at
$m = 25$ it recovers faster on CIFAR-20 as well. At $m=1$ the probe itself
degrades the retrained model, whose retain accuracy after 50 steps is
\numRelearnGoldDrStepfiftyCtenMone\%, \numRelearnGoldDrStepfiftyCtwentyMone\% and
\numRelearnGoldDrStepfiftyChundredMone\%, so we draw no conclusion from $m=1$.

\begin{figure}[t]
\begin{center}
\includegraphics[width=0.7\linewidth]{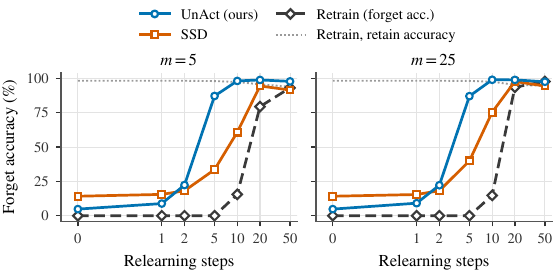}
\end{center}
\caption{Relearning on ViT-B/16 (CIFAR-10), learning rate $5\times10^{-4}$,
$m$ forget images per step mixed $4{:}1$ with retain images, evaluated in
bfloat16, with retain accuracy on 2{,}000 retain test images and forget accuracy
on the full forget test split. Mean over the five forget classes. UnAct's edit is undone within a few steps, faster than
SSD's; the retrained model stays at zero forget accuracy for the first five
steps.}
\label{fig:vit-relearn}
\end{figure}

On ViT-B/16 we use a learning rate of $5\times10^{-4}$, at which the retrained
model's retain accuracy stays high (Figure~\ref{fig:vit-relearn}). After five
steps at $m=5$, UnAct's forget accuracy is back to
\numVitRelearnDfUnActMfiveSfive\%, against \numVitRelearnDfSSDMfiveSfive\% for SSD
and \numVitRelearnDfGoldMfiveSfive\% for the retrained model, averaged over the
five forget classes.

\section{Coverage}
\label{app:coverage}

If a round selects $m$ of the scope's $N$ units, the fraction of the scope edited
after $k$ rounds satisfies the elementary union bound
\begin{equation}
\Gamma_k \;=\; \frac{1}{N}\Bigl|\bigcup_{t=1}^{k} \mathcal{M}_t\Bigr|
\;\leq\; \min\!\Bigl(1,\; \frac{k\,m}{N}\Bigr),
\label{eq:coverage}
\end{equation}
with equality when successive rounds select disjoint sets. The bound itself is
trivial. The empirical point is that measured coverage stays close to it for
small $k$, so the percentile $p$ understates the size of the edit by up to a
factor of $k$.

\begin{figure}[t]
\begin{center}
\includegraphics[width=\linewidth]{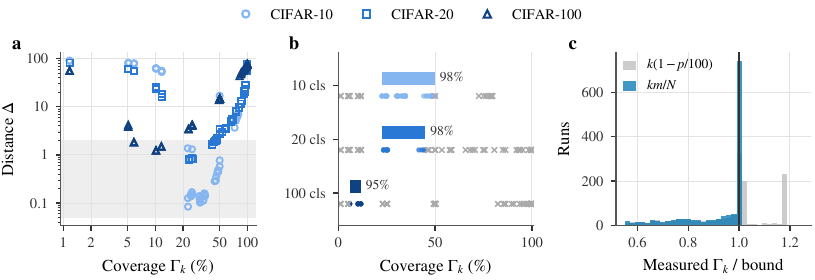}
\end{center}
\caption{Cumulative coverage $\Gamma_k$ organises UnAct's hyperparameters.
(a) Across the grid at $\gamma \leq 0.1$, $\Delta$ is U-shaped in coverage.
(b) A coverage window per dataset (bar) against configurations with $\Delta \leq
2$ (dots) and above (crosses); the percentage is the share of configurations the
window classifies correctly. The windows were fitted on this grid, so they
describe it rather than forecast a held-out one. (c) Measured coverage divided by
two bounds: the naive $\min(1, k(1-p/100))$ is exceeded in
\numCoverageExceedNaive{} of \numCoverageRuns{} runs, because selection uses
$\geq$ against an interpolated percentile and so keeps slightly more than a
$(1-p/100)$ share of units, while the bound $km/N$ of Eq.~\eqref{eq:coverage} is
exceeded in \numCoverageExceedKmN{} runs.}
\label{fig:coverage}
\end{figure}

Figure~\ref{fig:coverage} measures the coverage $\Gamma_k$ of
Eq.~\eqref{eq:coverage} over UnAct's whole grid. The bound holds in every run, and
the usable configurations occupy a band of coverage that moves to smaller values
as the number of classes grows.
\end{document}